%% file: main.tex
\documentclass{article} 
\usepackage{iclr2025_conference,times}

\input{math_commands.tex}

\usepackage{tikz}
\usetikzlibrary{arrows.meta, positioning, calc}
\usepackage{subcaption}
\usepackage{graphicx,adjustbox}

\usepackage{hyperref}
\usepackage{url}
\usepackage{graphicx}
\usepackage{amsmath}
\usepackage{tikz}
\usepackage{enumitem} 

\usepackage{graphicx}
\usepackage{subcaption}
\usepackage{float} 
\usepackage{amsmath} 
\usepackage{listings} 
\usepackage{xcolor} 
\usepackage{graphicx,array,booktabs}
\usepackage{amsthm}
\newtheorem{assumption}{Assumption}
\newtheorem{theorem}{Theorem}
\newtheorem{corollary}{Corollary}
\newtheorem{lemma}{Lemma}
\newtheorem{remark}{Remark}
\usepackage{amsmath, amssymb}   
\usepackage{bm}                

\usepackage{algorithm}         
\usepackage[noend]{algpseudocode} 
\usepackage{siunitx}

\usepackage{graphicx,adjustbox,xparse}

\definecolor{codegreen}{rgb}{0,0.6,0}
\definecolor{codegray}{rgb}{0.5,0.5,0.5}
\definecolor{codepurple}{rgb}{0.58,0,0.82}
\definecolor{backcolour}{rgb}{0.95,0.95,0.92}

\lstdefinestyle{mystyle}{
    backgroundcolor=\color{backcolour},   
    commentstyle=\color{codegreen},
    keywordstyle=\color{blue},
    numberstyle=\tiny\color{codegray},
    stringstyle=\color{codepurple},
    basicstyle=\ttfamily\footnotesize,
    breakatwhitespace=false,         
    breaklines=true,                 
    captionpos=b,                    
    keepspaces=true,                 
    numbers=left,                    
    numbersep=5pt,                  
    showspaces=false,                
    showstringspaces=false,
    showtabs=false,                  
    tabsize=2
}
\usepackage{siunitx} 

\usepackage{amsthm}

\newtheorem{proposition}{Proposition}

\newcommand{\order}{\mathcal{O}}
\newcommand{\TV}{\mathrm{TV}}
\newcommand{\dd}{\mathrm{d}}
\usepackage{tikz}
\usetikzlibrary{arrows.meta,positioning,fit,shapes.multipart}
\title{Diffusion Models are Molecular Dynamics Simulators}

\author{Justin S. Diamond \& Markus Lill \\
Department of Pharmaceutical Sciences Science\\
University of Basel\\
Basel, 4058, Switzerland \\
\texttt{justin@hetzerk.com} \\
}

\iclrfinalcopy 
\usepackage{fancyhdr}
\usepackage{fvextra} 
\DefineVerbatimEnvironment{code}{Verbatim}{
  breaklines=true,      
  breakanywhere=true,   
  fontsize=\small,      
  tabsize=2,            
}
\begin{document}

\maketitle

\begin{abstract}
We show that a denoising–diffusion sampler equipped with a \emph{harmonic adapter}---a quadratic coupling with an offset that ties neighbouring reverse‑time iterates---is exactly an Euler–Maruyama (EM) integrator for overdamped Langevin dynamics. One reverse step with spring stiffness $k$ integrates the SDE with implicit step size $\Delta t=\beta/(2k)$; the drift is the learned score (or energy gradient). This identity reframes molecular dynamics (MD) through diffusion: the accuracy of trajectories and equilibrium statistics is controlled by two dials, \emph{model capacity} (via universal approximation of the drift) and the \emph{number of denoising steps} $N$, rather than by a fixed, tiny MD timestep. 

The practical consequence is a data‑driven MD framework that learns \emph{forces for free} from i.i.d.\ configurations, requires no hand‑crafted force fields, and can be run with a small, distillable number of reverse steps while preserving the Boltzmann law of the learned energy. We prove pathwise KL bounds that separate discretisation error $\mathcal O(\sum_n \Delta t^2)$ from score error, show how temperature enters through the spring, and demonstrate MD‑like temporal correlations on molecular trajectories generated entirely by a score model trained on static samples.
\end{abstract}

\section{Introduction}

Diffusion models have reshaped generative modeling, from vision to molecular data~\cite{sohl2015deep, ho2020denoising, song2020score}, delivering high-fidelity samples and robust likelihood surrogates~\cite{karras2022elucidating}. In molecular modeling, however, the objective is not merely to match static distributions but to generate configurations that are consistent with physical constraints and thermodynamics~\cite{noe2019boltzmann}. Classical approaches such as Langevin Molecular Dynamics (LMD)~\cite{ermak1978brownian, leimkuhler2015molecular} integrate stochastic differential equations to evolve systems in time, enabling exploration of conformational landscapes and analysis of dynamical processes~\cite{schlick2010molecular}.

A central difficulty shared by both diffusion samplers and LMD is sampling from complex, high-dimensional \emph{non-iid} target distributions. Molecular interactions (covalent bonding, electrostatics, van der Waals forces) induce long-range and multi-scale correlations that invalidate iid assumptions and complicate both inference and analysis~\cite{frenkel2001understanding}. Recent diffusion models specialized to molecules seek to respect these structures~\cite{xu2022geodiff, anand2022protein}, including symmetry-equivariant formulations that help preserve physical invariances~\cite{hoogeboom2022equivariant}. Yet ensuring that generated samples obey physical laws and exhibit meaningful temporal statistics remains challenging.

Temporal correlations are especially important for scientific use cases. Numerical solvers for ODE/SDEs produce trajectories via sequential updates that encode short-time physics and enable computation of path-dependent observables~\cite{hairer2006geometric}. Such trajectories underpin studies of folding kinetics and reaction mechanisms~\cite{dill2012protein}. By contrast, standard diffusion sampling treats reverse-time updates as conditionally independent draws, which yields excellent equilibria but discards dynamical structure.

Bridging these views motivates introducing \emph{structured interactions} into the reverse process. In diffusion models, non-iid noise or couplings can emulate dependencies analogous to those in MD; however, analyzing the correctness and convergence of such modifications is subtle~\cite{song2021maximum}. Our perspective is to endow reverse-time iterations with a simple, \emph{harmonic adapter}---a quadratic coupling with an offset that ties neighboring reverse steps---so that each denoising update admits a physically interpretable meaning.

\paragraph{Key idea (in one line).} We show that a denoising--diffusion sampler equipped with a harmonic adapter performs, step-for-step, the Euler--Maruyama (EM) update for overdamped Langevin dynamics at an \emph{implicit} resolution $\Delta t = \beta/(2k)$ determined by the spring stiffness $k$. In effect, the denoising grid becomes a controllable \emph{resolution dial} for MD-like trajectories, while retaining the probabilistic benefits of diffusion.

\paragraph{Why this matters (computational pain point).} Molecular scientists routinely need trajectories spanning micro- to millisecond scales to compute kinetics, nonequilibrium responses, and transport properties. A standard LMD integrator with a $2\,\mathrm{fs}$ step requires ${\sim}10^{\!9}$ updates to reach $2\,\mu\mathrm{s}$ of simulated time, often translating into days of wall-clock GPU time even with aggressive parallelization. In contrast, score-based diffusion can produce equilibrium-quality conformations in \emph{tens} of reverse steps after distillation, suggesting a path to $10^2$--$10^3\times$ reductions in wall-clock cost without abandoning statistical exactness in the learned energy.

\paragraph{Contributions.} 
\textbf{(1)} We provide an exact algebraic identity: one reverse denoising step with a harmonic adapter equals one EM step for overdamped Langevin, with drift given by the learned score/energy gradient and noise given by the EM Gaussian. 
\textbf{(2)} This identity yields a clean accuracy budget that separates model (score) error from grid (reverse-schedule) error, and supports principled temperature/variance control during inference. 
\textbf{(3)} The same quadratic coupling enables fully time-parallel execution: batches index trajectory slices that can be updated simultaneously while preserving the correct local-in-time law. 
\textbf{(4)} Empirically, glued trajectories reproduce ensemble statistics and exhibit MD-like temporal correlations on the learned energy landscape on open-source pretrained models, enabling trajectory-level observables at diffusion costs.

\paragraph{Outlook.} Treating reverse-time diffusion as an MD integrator on the learned potential opens a route to \emph{trajectory} generators with physical semantics, modular coupling to MCMC and enhanced sampling, and higher-order adapters---all while retaining the scalability and training simplicity that make diffusion attractive in the first place.

\subsection*{From trajectories to equilibria—and back again}

A diffusion model with a simple \emph{quadratic (harmonic) coupling} between consecutive reverse‑time iterates is an MD integrator in disguise. Section~\ref{sec:score-model-implications} formalises this: completing the square in the exact EM transition shows that the reverse kernel of denoising–diffusion with the harmonic offset coincides with the EM kernel
\[
x_{n+1}=x_n-\Delta t\,\nabla V_\theta(x_n)+\sqrt{2D\,\Delta t}\,\xi_n,
\qquad 
\Delta t=\frac{\beta}{2k}.
\]
Here $\nabla V_\theta$ is the score/energy gradient learned from i.i.d.\ data. In this reframing, \emph{time resolution is no longer a hard bottleneck}: fidelity is governed by (a) how well the network approximates the true drift (universal approximation) and (b) how many reverse steps we choose (which we can distil to a few dozen without retraining the score).

Standard Langevin MD advances at femtosecond timesteps to remain stable; reaching microseconds requires ${\sim}10^6$ steps. Diffusion samplers, by contrast, operate on a coarse reverse‑time grid whose size we can set for inference. The harmonic adapter makes this more than an analogy: one reverse step \emph{is} one EM step, with an \emph{implicit} $\Delta t$ encoded by the spring. Thus, long‑time, MD‑like trajectories (and time‑correlated observables such as dihedral autocorrelations) can be generated with wall‑clock cost proportional to the \emph{number of denoising iterations}, not the micro‑timestep count of MD.

Training requires only static, i.i.d.\ configurations. A denoising objective learns the score $\nabla\log p$; we view $E_\theta:=-\log p_\theta$ as a \emph{learned energy} whose gradient plays the role of a force field. No bespoke bonded/non‑bonded decomposition or parameter fitting is needed. The resulting sampler integrates the learned energy landscape with EM accuracy, and a one‑step Metropolis check (optional) makes the terminal law exactly Boltzmann in $E_\theta$.

The adapter penalises the \emph{mismatch} $x_{n+1}-x_n+\Delta t\,\nabla V_\theta(x_n)$ with stiffness $k=\beta/(2\Delta t)$. Because the mismatch is $\mathcal O(\sqrt{\Delta t})$, the quadratic energy contributes only $\mathcal O(\Delta t)$ to the exponent; the continuum limit is well behaved. Operationally, this turns independent denoising updates into a \emph{time‑correlated} trajectory consistent with Langevin physics.

Accuracy separates cleanly into \emph{score error} and \emph{grid error}. Our pathwise bound (Sec.~\ref{sec:main-md-limit}) shows
\[
\KL\!\big(\mathcal L(\tilde X_{[0,T]})\,\|\,\mathcal L(X_{[0,T]})\big)
\;\le\; T\,\bar\varepsilon^2\;+\; C\sum_{n=0}^{N-1}\Delta t^2.
\]
so better networks (universal approximation) and fewer but suitably chosen reverse steps $N$ jointly control fidelity. In practice, distilled schedules with $N\!\in\![10,50]$ already reproduce MD‑like autocorrelations while costing orders of magnitude less than conventional MD.

This viewpoint recasts “force fields” as \emph{learned scores} trained on i.i.d.\ data and turns MD into \emph{generative inference}. Temperature enters only via the spring ($k=\beta/(2\Delta t)$), enabling explicit control of effective temperature and correlation length at inference, and opening the door to time‑parallel sampling and easy coupling to other stochastic algorithms (MCMC, metadynamics, alchemy) without redesigning the network.

The rest of the paper is organized as follows: we provide background on diffusion models and present our theoretical framework showing equivalence between diffusion models with harmonic bias and LMD. We discuss the implications for parallelization across time steps and the potential for efficient simulation of molecular systems. In the appendix we discuss in depth the equivalence, illustrate results related to the underdamped langevin diffusion, and discuss future work about using this formalism to do MCMC, Metadyanmics, and Alchemical Free Energy estimations in parallel.

\begin{figure}[ht!]
    \centering
    \begin{tikzpicture}
        \node at (0,0) {\includegraphics[width=0.4\textwidth]{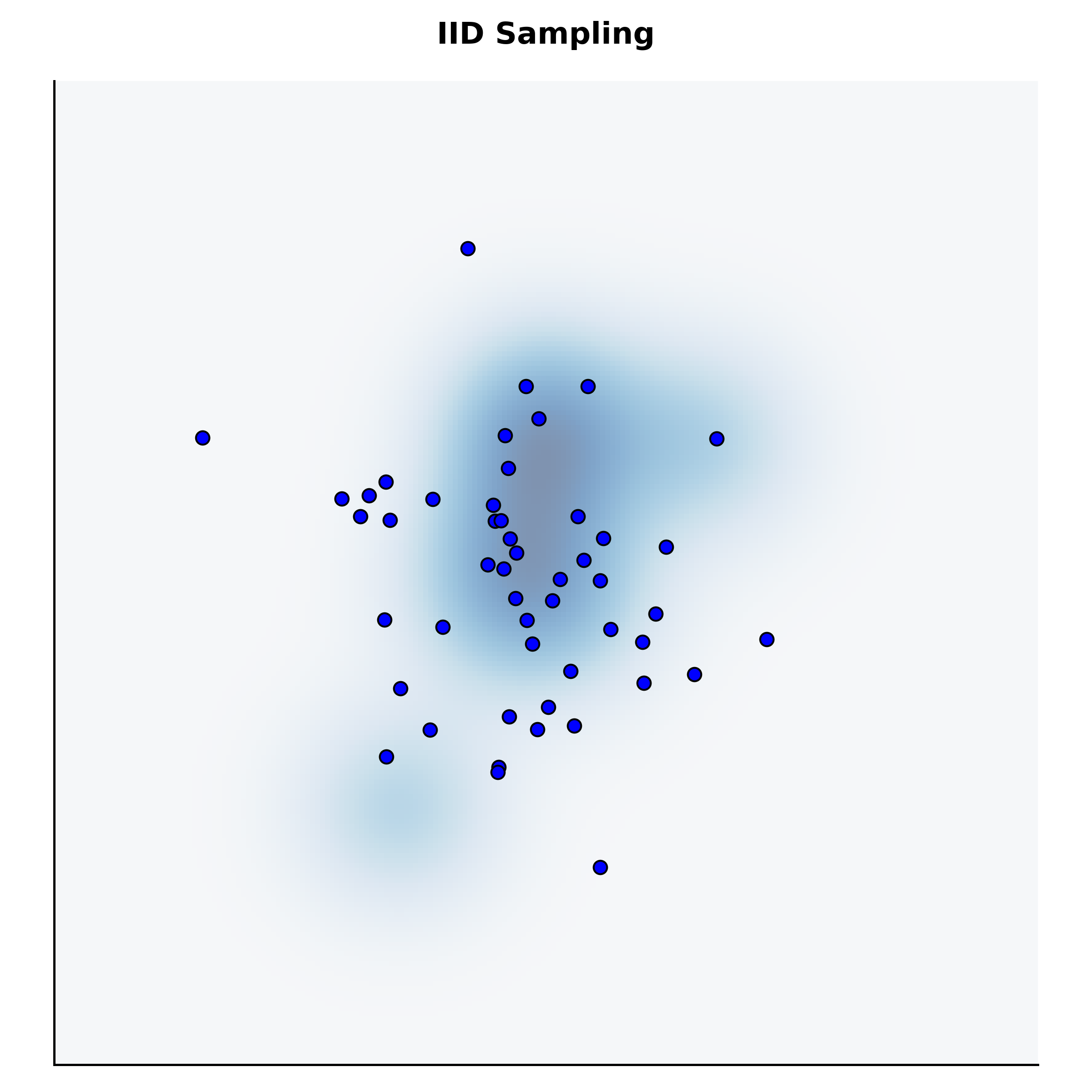}};
        \node at (7,0) {\includegraphics[width=0.4\textwidth]{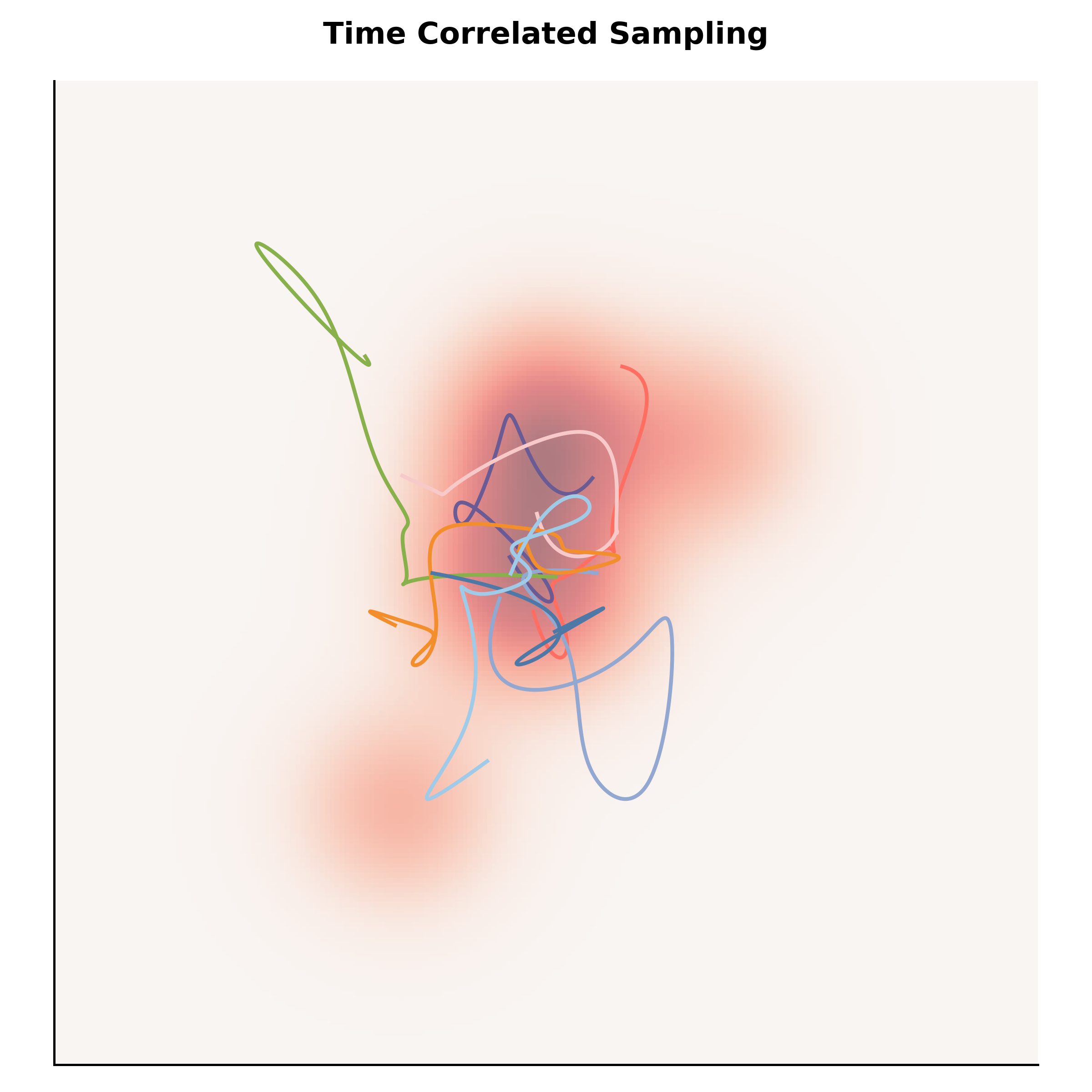}};
        \draw[thick,->] (3,0) -- (4,0);
    \end{tikzpicture}
    \caption{Left: traditional i.i.d.\ diffusion sampling ignores temporal structure.  Right: our harmonically‑coupled sampler recovers time‑correlated trajectories consistent with MD.}
\end{figure}
\section{Definitions, Notation, and What Is Proven}
\subsection{Over‑damped Langevin Dynamics (continuous time)}
\label{sec:langevin-cont}
\textbf{Units and constants.}
Throughout we choose \emph{friction units} such that
\[
   \gamma = 1
   \quad\Longrightarrow\quad
   D = \frac{k_{\!B}T}{\gamma}=k_{\!B}T,
   \quad
   \beta D \;=\; 1.
\]
Stating this explicitly prevents hidden factors of~$\gamma$
or~$k_{\!B}$ from re‑appearing later.

A single particle in a conservative potential
\(V:\mathbb{R}^{d}\!\to\!\mathbb{R}\) at temperature~\(T\) therefore
obeys
\begin{equation}\label{eq:langevin-sde}
   d\bm{x}(t)
   = -\nabla V\bigl(\bm{x}(t)\bigr)\,dt
     +\sqrt{2D}\,d\bm{w}(t),\qquad
   D=k_{\!B}T,\; \beta\!=\!1/D .
\end{equation}
\subsection{Euler–Maruyama time discretisation}
\label{sec:em}

Discretising the overdamped Langevin SDE~\eqref{eq:langevin-sde}
with a fixed step size \(\Delta t>0\) gives the classical
Euler–Maruyama (EM) integrator
\begin{equation}\label{eq:em-step}
   \mathbf x_{n+1}
   = \mathbf x_n - \nabla V(\mathbf x_n)\,\Delta t
     + \sqrt{2D\,\Delta t}\;\boldsymbol\xi_n,
   \qquad
   \boldsymbol\xi_n\sim\mathcal N(\mathbf 0,\mathbf I_d).
\end{equation}
Hence, **conditioned on \(\mathbf x_n\)**, the next point
\(\mathbf x_{n+1}\) is Gaussian
\begin{equation}\label{eq:em-gauss}
   p(\mathbf x_{n+1}\!\mid\mathbf x_n)
   = \frac{%
        \exp\!\bigl[
          -\tfrac{\beta}{4D\Delta t}
          \,\|\mathbf x_{n+1}-\mathbf x_n
                 +\Delta t\,\nabla V(\mathbf x_n)\|^{2}
        \bigr]}%
        {(4\pi D\Delta t)^{d/2}}\; .
\end{equation}
This is the **exact** EM transition kernel; no other terms appear in
the exponent.

\subsection{Harmonic reformulation (“quadratic glue”)}
\label{sec:harmonic-glue}

For later use it is convenient to rewrite
\eqref{eq:em-gauss} as a Boltzmann factor with a
\emph{quadratic coupling} (the \emph{harmonic adapter})
\begin{align}
   p(\mathbf x_{n+1}\!\mid\mathbf x_n)
   &= Z(\mathbf x_n)^{-1}\;
      \exp\!\Bigl[
         -\beta\,
         \frac{k(\Delta t)}{2}\;
         \bigl\|\mathbf x_{n+1}-\mathbf x_n
                +\nabla V(\mathbf x_n)\Delta t\bigr\|^{2}
      \Bigr],
      \label{eq:em-boltz} \\
   k(\Delta t) &= \frac{1}{2D\Delta t}.           \label{eq:k-dt}
\end{align}
A key observation is that the \emph{mismatch}
\(
\mathbf x_{n+1}-\mathbf x_n+\nabla V(\mathbf x_n)\Delta t
= \mathcal O(\sqrt{\Delta t})
\),
so the spring energy contributes only \(\mathcal O(\Delta t)\) to the
exponent even though the stiffness
\(k(\Delta t)\sim\Delta t^{-1}\) diverges.
Consequently the limit \(\Delta t\to0\) remains well behaved and the
kernel \eqref{eq:em-boltz} converges to the continuum Langevin
propagator, \S\ref{sec:harm-adapter}.

\section{Implications for Score‑Based (Denoising–Diffusion) Models}
\label{sec:score-model-implications}

\subsection{From IID denoising to molecular dynamics: the equivalence principle}
\label{sec:equivalence-principle}
\textbf{Key message.} With the harmonic (quadratic) adapter, each reverse‑diffusion step is
\emph{exactly} an Euler–Maruyama (EM) step for overdamped Langevin with step \(\Delta t\). Thus, if the score/force proxy equals the true force and the denoising grid is refined, the glued diffusion paths converge to MD trajectories targeting the same Boltzmann law. The spring \(k\) is not a heuristic: it \emph{is} the EM stiffness \(k(\Delta t)=\frac{1}{2D\Delta t}=\frac{\beta}{2\Delta t}\) (Eq. (10), *p. 6*). In practice, this gives a clean mapping between diffusion time and MD time and explains the “resolution dial” observed in the figures (e.g., Fig. 1 on *p. 3*, Figs. 7–9 on *p. 11*).  

We use friction units \(\beta D=1\). Let \(V:\R^d\to\R\) be the physical potential (unknown or known), and\[
\pi_\beta(dx) \propto e^{-\beta V(x)}\,dx
\] the target density.

\subsection{IID baseline: training objects and learned energy}
\label{sec:iid-baseline}
Let \(p_t\) be the data density after IID Gaussian corruption with variance \(\sigma_t^2=2Dt\). A standard denoising objective fits the \emph{time‑dependent score}
\[
s^{\mathrm{iid}}_t(x)\;\approx\;\nabla_x\log p_t(x).
\]
For notational clarity we also define a \emph{learned energy}
\begin{equation}\label{eq:learned-energy}
  E_\theta(x):=-\log p_\theta(x)+\mathrm{const},\qquad \nabla E_\theta(x)=-s^{\mathrm{iid}}_0(x).
\end{equation}
Either \(s^{\mathrm{iid}}_{t_n}\) (time‑dependent) or \(\nabla E_\theta\) (time‑0) can be used at inference as a drift proxy \(g_n\).

\paragraph{Universal approximability of the force/score (assumption → guarantee).}
On compact sets \(\mathcal K\subset\R^d\), standard universal approximation results imply that neural nets can approximate smooth maps and their gradients uniformly. We capture this as a modeling hypothesis:
\begin{assumption}[Universal approximation of the drift]\label{ass:ua}
There exists a sequence \(g^{(m)}\) (e.g., scores/energies with increasing capacity or better training) such that
\(
\sup_{x\in\mathcal K}\|\,g^{(m)}(x)-\nabla V(x)\|\to 0
\)
as \(m\to\infty\).
\end{assumption}
Assumption~\ref{ass:ua} is the bridge that lets the learned drift substitute for the physical force in Theorem~\ref{thm:md-limit} below.

\subsection{Inference with quadratic glue (harmonic adapter)}
\label{sec:harmonic-inference}
Fix a denoising step \(\Delta t>0\) and set
\begin{equation}\label{eq:harmonic-proposal}
  m_n \;:=\; x_n - D\Delta t\, g_n(x_n),
  \qquad
  k(\Delta t)=\frac{1}{2D\Delta t}=\frac{\beta}{2\Delta t}.
\end{equation}
\emph{Conditioned on \(x_n\)}, the next state is drawn from the \textbf{Gaussian} kernel
\begin{equation}\label{eq:harmonic-kernel}
  p(x_{n+1}\!\mid x_n)
  \;=\; \frac{\exp\!\big[-\frac{\beta}{4D\Delta t}\,\|x_{n+1}-m_n\|^2\big]}{(4\pi D\Delta t)^{d/2}}.
\end{equation}
This is precisely the EM kernel with drift \(-g_n\) (compare the quadratic‑glue Boltzmann form in §\ref{sec:harmonic-glue}). The algebraic identity \(k(\Delta t)=\beta/(2\Delta t)\) makes \(\Delta t\) an \emph{implicit} resolution scale controlled by the spring (Eq. (10), *p. 6*). 

\paragraph{Optional exactness for the learned energy.}
Appending one Metropolis–Hastings accept/reject step with target \(\propto e^{-\beta E_\theta}\) makes the terminal law exactly Boltzmann on \(E_\theta\), with acceptance \(1-\mathcal O(\Delta t)\). In practice we omit MH for speed unless exact stationarity w.r.t.\ \(E_\theta\) is required.

\subsection{Main equivalence and limiting theorem (diffusion $\Rightarrow$ MD)}
\label{sec:main-md-limit}
We now state the finite‑grid bound and its limiting consequence. Let \(X_t\) solve the overdamped SDE
\(
\mathrm dX_t=-\nabla V(X_t)\mathrm dt+\sqrt{2D}\,\mathrm dW_t
\)
and let \(\tilde X_t\) be the piecewise‑constant‑drift interpolation of the glued sampler \eqref{eq:harmonic-kernel} on \([0,T]\) (same Brownian path). Suppose the inference drift satisfies a uniform error bound
\(
g_n=\nabla V+\varepsilon_n,\ \ \sup_{n,x}\|\varepsilon_n(x)\|\le \bar\varepsilon.
\)

\begin{theorem}[Finite‑schedule pathwise KL bound]\label{thm:path-kl-imp}
With \(\beta D=1\), there is a constant \(C=C(D,L)\) depending on the Lipschitz constant \(L\) of \(\nabla V\) such that
\[
\KL\!\big(\mathcal L(\tilde X_{[0,T]})\,\|\,\mathcal L(X_{[0,T]})\big)
\;\le\; T\,\bar\varepsilon^2\;+\; C\sum_{n=0}^{N-1}\Delta t^2.
\]
Consequently, by Pinsker,
\(
\|\mathcal L(\tilde X_{[0,T]})-\mathcal L(X_{[0,T]})\|_{\mathrm{TV}}
\le \tfrac12\sqrt{T\,\bar\varepsilon^2+C\sum_n\Delta t^2}.
\)
\end{theorem}

\begin{proof}[Derivation sketch]
On each interval \([t_n,t_{n+1})\), the glued drift equals \(-g_n(x_n)\). The exact drift is \(-\nabla V(\tilde X_t)\). Girsanov gives
\(
\KL=\frac{1}{4D}\E\!\int_0^T\!\| -g_n(\tilde X_{t_n})+\nabla V(\tilde X_t)\|^2\mathrm dt.
\)
Split the gap into model and Lipschitz parts:
\(
-\varepsilon_n(\tilde X_{t_n})+(\nabla V(\tilde X_{t_n})-\nabla V(\tilde X_t)).
\)
The first contributes \(\tfrac{1}{4D}\bar\varepsilon^2\Delta t\) per interval. The second is bounded by \(L^2\E\|\tilde X_t-\tilde X_{t_n}\|^2\), and EM moment bounds give \(\E\|\tilde X_t-\tilde X_{t_n}\|^2\le C'(t-t_n)\). Integrate to obtain \(C\Delta t^2\) per interval and sum. \emph{QED}.  \,\,\,\,\,\,\,\,\,\,\,\,\,\,\,\,\,\,\,\,\,\,\,\,\,\,\,\,\,\,\,\,\,\,\,\,\,\,\,\,\,\,\,\,\,\,\,\,\,\,\,\,\,\,\,\,\,\,\,\,\,\,\,
\end{proof}

\begin{theorem}[Diffusion\,$\Rightarrow$\,MD in the fine‑grid/universal‑approximation limit]\label{thm:md-limit}
Let \(T>0\) be fixed. Let \(\Delta t_N\to 0\) with \(N\Delta t_N\to T\), and let \(g^{(m)}\) satisfy Assumption~\ref{ass:ua} on compacts. Then there exist sequences \(N\to\infty\) and \(m\to\infty\) (refining grid and model) such that
\[
\KL\!\big(\mathcal L(\tilde X^{(N,m)}_{[0,T]})\,\|\,\mathcal L(X_{[0,T]})\big)\to 0,\qquad
\|\mathcal L(\tilde X^{(N,m)}_{[0,T]})-\mathcal L(X_{[0,T]})\|_{\mathrm{TV}}\to 0.
\]
Equivalently, for every bounded Lipschitz path functional \(F\),
\(
|\E F(\tilde X^{(N,m)}_{\cdot})-\E F(X_{\cdot})|\to 0.
\)
\end{theorem}

\begin{proof}
Apply Theorem~\ref{thm:path-kl-imp} with \(\bar\varepsilon=\bar\varepsilon(m)\to 0\) (Assumption~\ref{ass:ua}) and \(\sum_n\Delta t_N^2\le T\max_n\Delta t_N\to 0\). Pinsker gives TV convergence; bounded‑Lipschitz convergence follows. \emph{QED}.
\end{proof}

 The harmonic‑glued diffusion sampler is an \emph{MD integrator} in disguise: in the fine‑grid/\emph{universal score} limit it reproduces overdamped Langevin on \(V\). In particular, time‑correlated observables (ACFs, Green–Kubo integrals), free‑energy path estimators, and equilibrium statistics computed from glued paths converge to their MD counterparts. This explains the empirical MD‑like trajectories and resolution/ESS effects observed in the experiments and figures (e.g., Figs. 10–17).

\textbf{Parallelism and speed.}
Each step is one Gaussian draw centered at \eqref{eq:harmonic-proposal}; all batch elements update independently. Combined with distillation to small \(N\), this yields orders‑of‑magnitude wall‑clock reductions for long effective times.

\subsection{Continuous‑time limit}
\label{sec:ct-limit}

With $D=\beta^{-1}$ the mean drift of
\eqref{eq:harmonic-kernel} reproduces, up to $o(\Delta t)$,
the reverse‑time SDE
\begin{equation}\label{eq:reverse-sde}
   d\mathbf x_t
   = -s^{\text{iid}}_t(\mathbf x_t)\,dt
     + \sqrt{2D}\,d\bar{\mathbf w}_t,
\end{equation}
so that, as $\Delta t\!\to\!0$,
the generated path converges weakly to continuous‑time Langevin dynamics
on the unknown potential $E_\theta$.

Because every update is a single Gaussian draw whose centre
\eqref{eq:harmonic-proposal} involves only a \emph{forward} call to the
score network, the wall‑clock cost is identical to that of the i.i.d.\
sampler—yet it yields trajectory data amenable to downstream molecular
analysis.
\subsection{Summary algorithm (pseudocode)}
\begin{algorithm}[H]
\caption{Batch Euler–Maruyama sampler with Metropolis correction}
\begin{algorithmic}[1]
\Require batch size $B$, total steps $T$, step size $\Delta t$,  
         score network $s^{\text{iid}}_{t} : \mathbb R^d \!\to\! \mathbb R^d$
\Statex

\State \textbf{Initialisation}
\For{$b = 0 \;\textbf{to}\; B$}
   \State Sample $\mathbf x_{0,b} \sim \mathcal N(\mathbf 0,\mathbf I_d)$
\EndFor
\Statex

\For{$t = 0 \;\textbf{to}\; T-1$}          \Comment{generation index}
  \For{$b = 0 \;\textbf{to}\; B$}          \Comment{batch index}
    \State $\mathbf m_{t,b} \gets
           \mathbf x_{t,b}
           - \Delta t\,s^{\text{iid}}_{t}(\mathbf x_{t,b})$
    \State Sample $\boldsymbol\xi_{t,b} \sim \mathcal N(\mathbf 0,\mathbf I_d)$
    \State $\mathbf x_{t+1,b} \gets
           \mathbf m_{t,b} + \sqrt{2D\Delta t}\,\boldsymbol\xi_{t,b}$
           \Comment{quadratic‑glue step}
    \If{Metropolis correction desired}
       \State Accept $\mathbf x_{t+1,b}$ with prob.\ $\alpha_{t,b}$ 
    \EndIf
  \EndFor
\EndFor
\end{algorithmic}
\end{algorithm}




\paragraph{Decoupling the forward noise schedule from the spring.}
The forward‑diffusion schedule \(\sigma_t\) is fixed during \emph{training}
and controls how strongly data are corrupted at each diffusion time
\(t\).  The harmonic spring \(k\) acts only \emph{at inference}.
Choosing a larger \(k\) (smaller implicit \(\Delta t\)) produces MD‑like,
highly correlated trajectories; choosing a smaller \(k\)
flattens correlations and mixes faster, at the cost of a slightly
larger \(O(\Delta t)\) bias that can be corrected with an
optional one‑step Metropolis check.



\paragraph{Conceptual distinction from conventional MD.}
\textbf{MD integrators} treat \(\Delta t\) as a hard‑coded parameter
        limited by stability (\(\Delta t\!\lesssim\!2\;\text{fs}\) for bonded
        hydrogens). \textbf{Harmonic‑glue diffusion} treats \(\Delta t\) as a
        \emph{derived} resolution scale.
        Stability is no longer a concern; larger \(\Delta t\) merely weakens
        the spring and shortens autocorrelation lengths. Thus, in our framework \(\Delta t\) is a \emph{resolution dial}, while the
primary user knob is the stiffness \(k\) that decorrelates samples at
constant bias order \(O(\Delta t)\).

Noise scheduling (training) and spring stiffness
(inference) serve \emph{orthogonal} purposes:
\(\sigma_t\) shapes the \emph{marginal} score prior, whereas \(k\)
sets the \emph{temporal resolution} of the generated trajectory via
\(\Delta t=\beta/2k\).  Making this separation explicit resolves the
apparent mismatch between MD time steps and reverse‑diffusion
iterations.
Additionally, higher-order methods \cite{kloeden1992numerical} emit other corresopnding batch neighboring relation as shown in Appendix C.

\begin{figure}[p]
  \centering
  \setlength{\tabcolsep}{6pt}
  \renewcommand{\arraystretch}{1.15}

  \caption{Radius‐of‐gyration traces for nine \(\mathrm{C}_{\!13}\)
           hydrocarbon conformers: diffusion sampler (centre) versus
           OpenMM reference (right).  Each plot is cropped by a
           molecule‐specific fraction to focus on the dynamic range of
           interest.}
  \label{fig:rg_grid}
  \vspace{0.7em}

  \begin{tabular}{@{}>{\centering\arraybackslash}m{0.22\linewidth}
                  >{\centering\arraybackslash}m{0.37\linewidth}
                  >{\centering\arraybackslash}m{0.37\linewidth}@{}}
    \toprule
    \textbf{Molecule} &
    \textbf{Diffusion \(R_g(t)\)} &
    \textbf{OpenMM \(R_g(t)\)} \\
    \midrule
    \includegraphics[width=\linewidth]{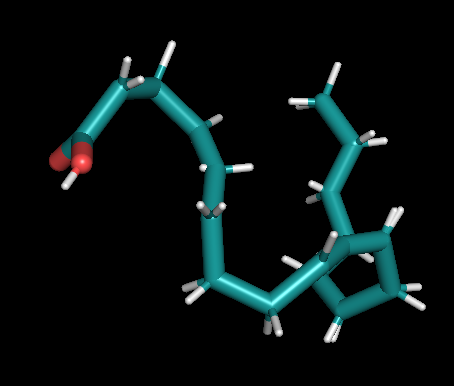} &
    \adjustbox{clip,trim=0 0 0 {0.095\height}}{%
      \includegraphics[width=\linewidth]{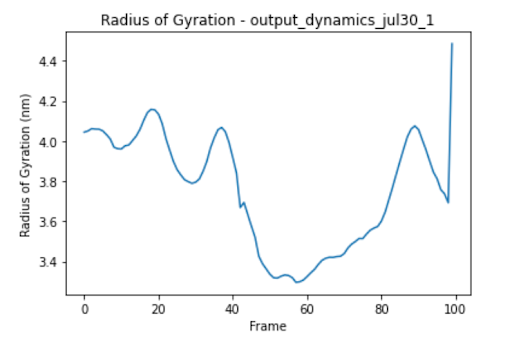}} &
    \adjustbox{clip,trim=0 0 0 {0.10\height}}{%
      \includegraphics[width=\linewidth]{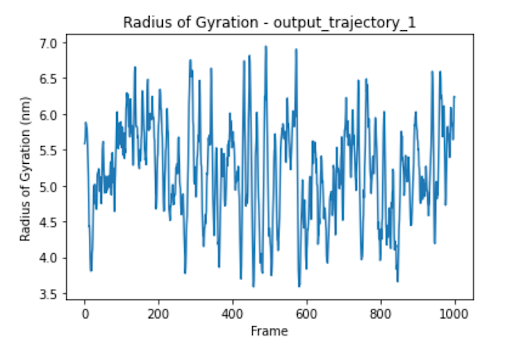}} \\[2pt]

    \includegraphics[width=\linewidth]{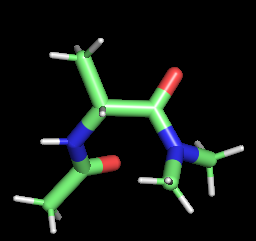} &
    \adjustbox{clip,trim=0 0 0 {0.1\height}}{%
      \includegraphics[width=\linewidth]{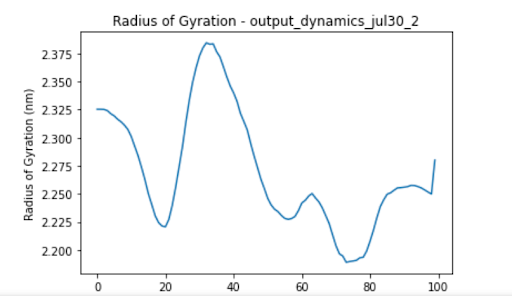}} &
    \adjustbox{clip,trim=0 0 0 {0.1\height}}{%
      \includegraphics[width=\linewidth]{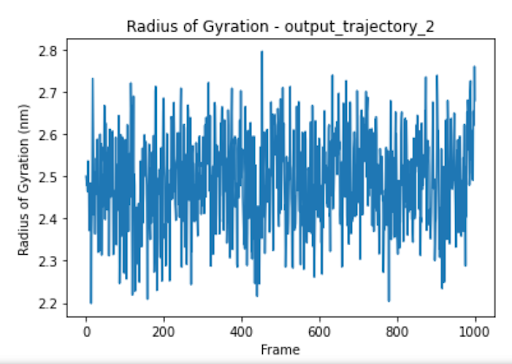}} \\[2pt]

    \includegraphics[width=\linewidth]{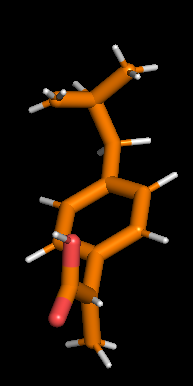} &
    \adjustbox{clip,trim=0 0 0 {0.1\height}}{%
      \includegraphics[width=\linewidth]{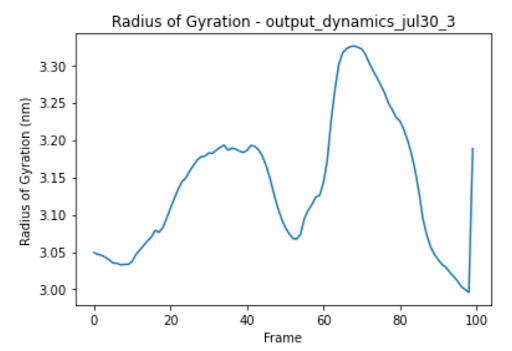}} &
    \adjustbox{clip,trim=0 0 0 {0.11\height}}{%
      \includegraphics[width=\linewidth]{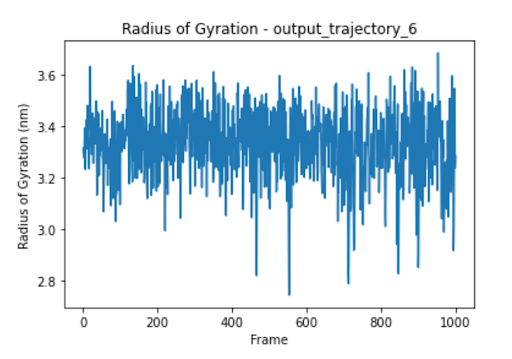}} \\[2pt]

    \includegraphics[width=\linewidth]{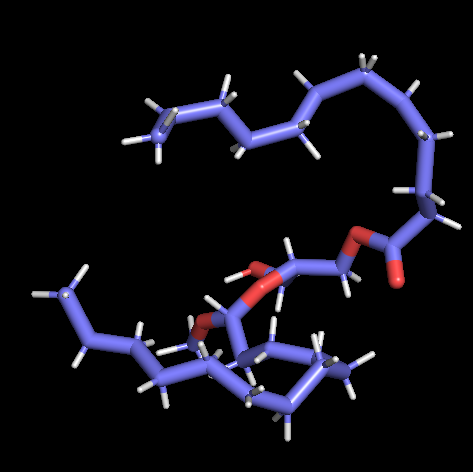} &
    \adjustbox{clip,trim=0 0 0 {0.09\height}}{%
      \includegraphics[width=\linewidth]{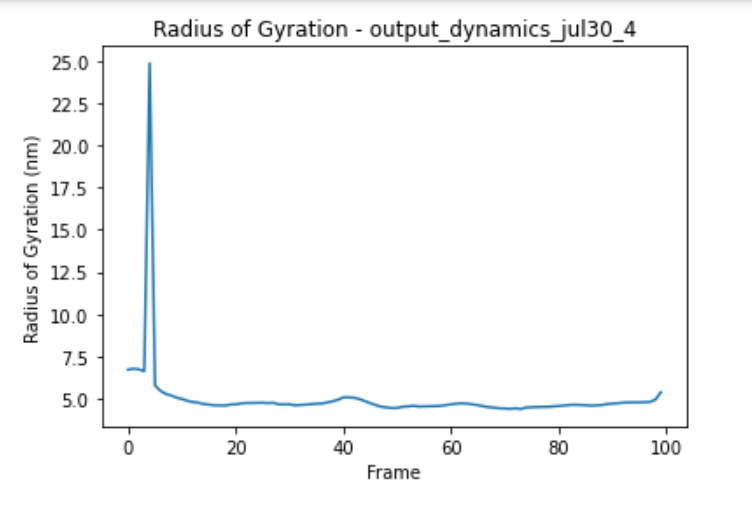}} &
    \adjustbox{clip,trim=0 0 0 {0.085\height}}{%
      \includegraphics[width=\linewidth]{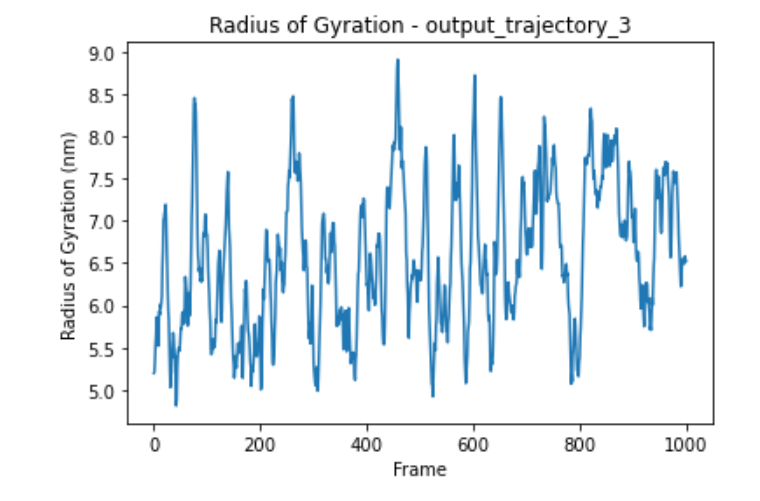}} \\[2pt]

    \includegraphics[width=\linewidth]{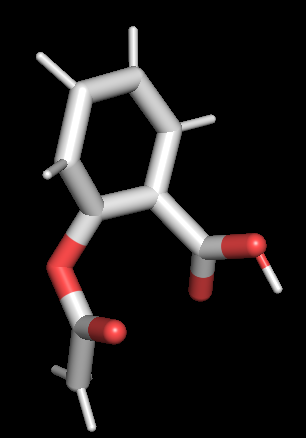} &
    \adjustbox{clip,trim=0 0 0 {0.1\height}}{%
      \includegraphics[width=\linewidth]{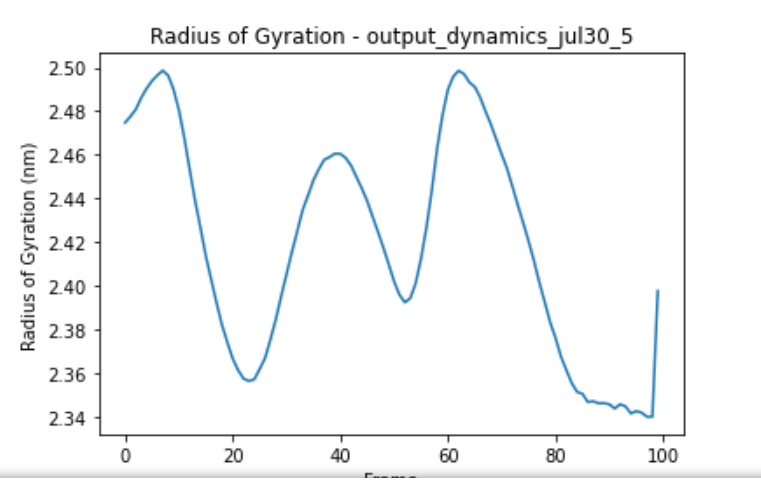}} &
    \adjustbox{clip,trim=0 0 0 {0.105\height}}{%
      \includegraphics[width=\linewidth]{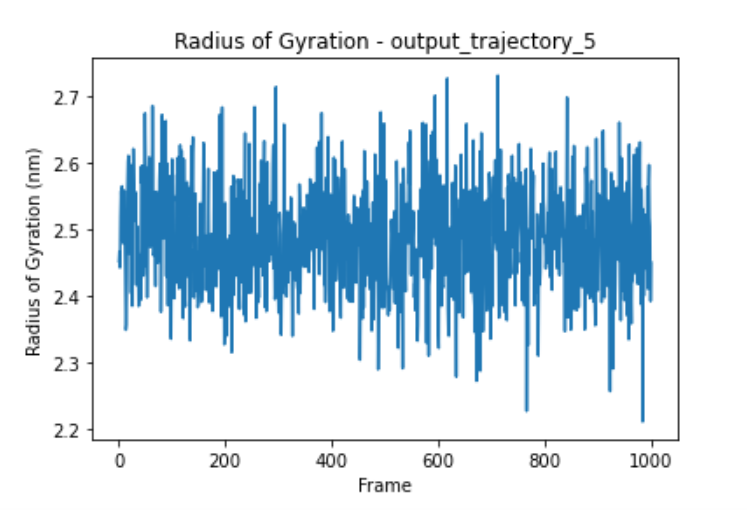}} \\[2pt]
    \bottomrule
  \end{tabular}
\end{figure}

\begin{figure}[p]\ContinuedFloat
  \centering
  \setlength{\tabcolsep}{6pt}
  \renewcommand{\arraystretch}{1.15}

  \begin{tabular}{@{}>{\centering\arraybackslash}m{0.22\linewidth}
                  >{\centering\arraybackslash}m{0.37\linewidth}
                  >{\centering\arraybackslash}m{0.37\linewidth}@{}}
    \toprule
    \textbf{Molecule} &
    \textbf{Diffusion \(R_g(t)\)} &
    \textbf{OpenMM \(R_g(t)\)} \\
    \midrule
    \includegraphics[width=\linewidth]{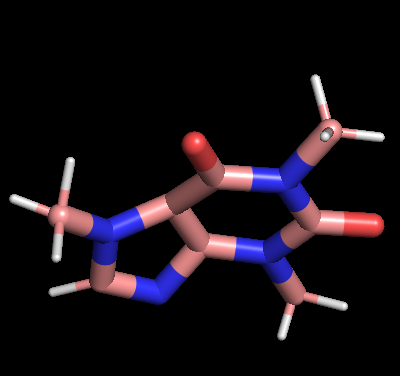} &
    \adjustbox{clip,trim=0 0 0 {0.067\height}}{%
      \includegraphics[width=\linewidth]{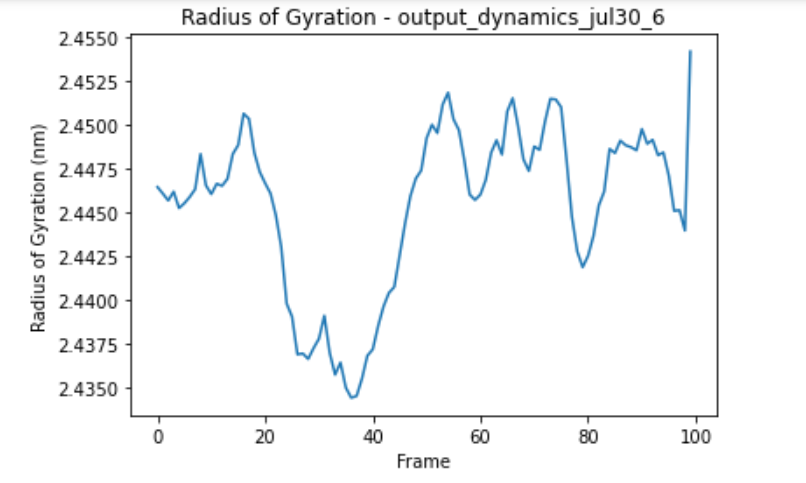}} &
    \adjustbox{clip,trim=0 0 0 {0.09\height}}{%
      \includegraphics[width=\linewidth]{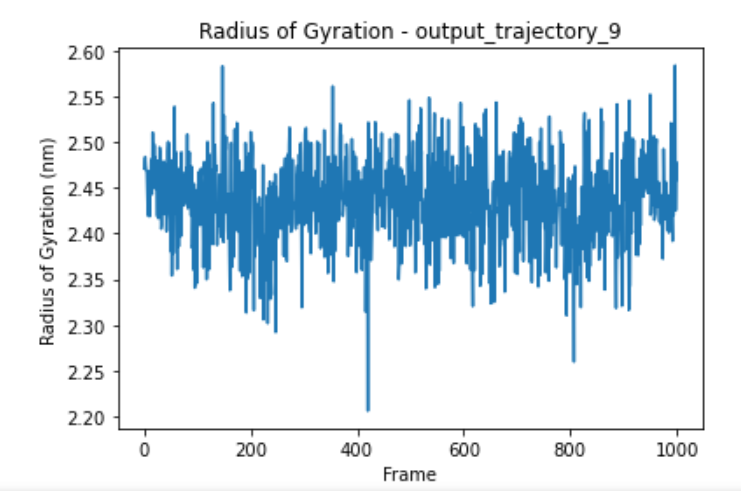}} \\[2pt]

    \includegraphics[width=\linewidth]{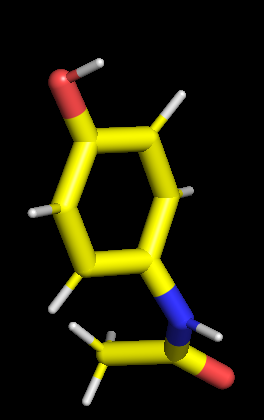} &
    \adjustbox{clip,trim=0 0 0 {0.10\height}}{%
      \includegraphics[width=\linewidth]{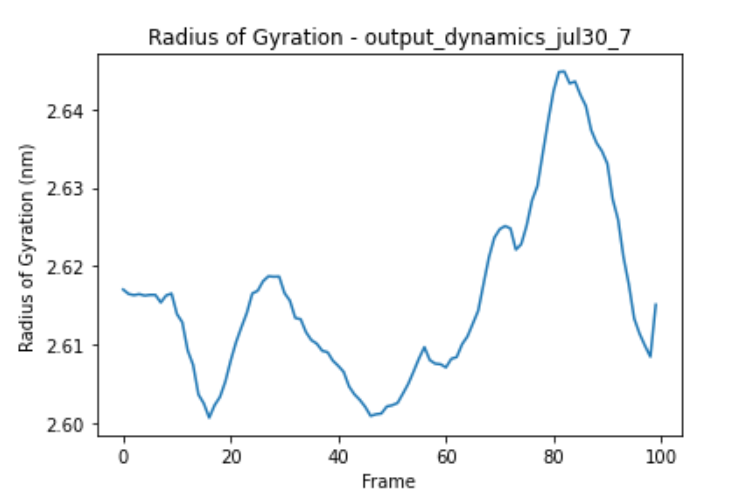}} &
    \adjustbox{clip,trim=0 0 0 {0.095\height}}{%
      \includegraphics[width=\linewidth]{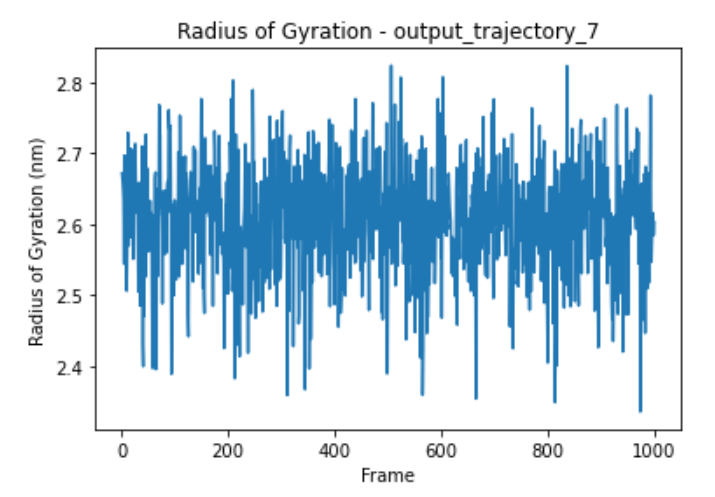}} \\[2pt]

    \includegraphics[width=\linewidth]{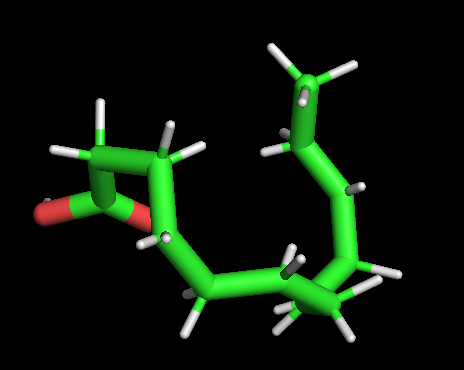} &
    \adjustbox{clip,trim=0 0 0 {0.08\height}}{%
      \includegraphics[width=\linewidth]{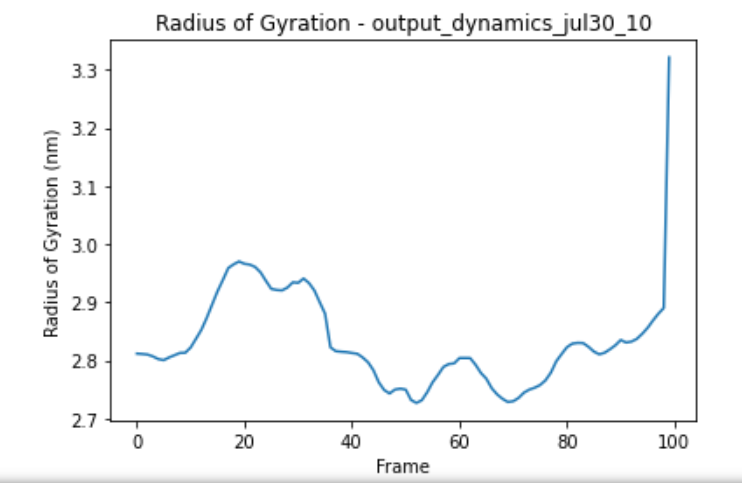}} &
    \adjustbox{clip,trim=0 0 0 {0.103\height}}{%
      \includegraphics[width=\linewidth]{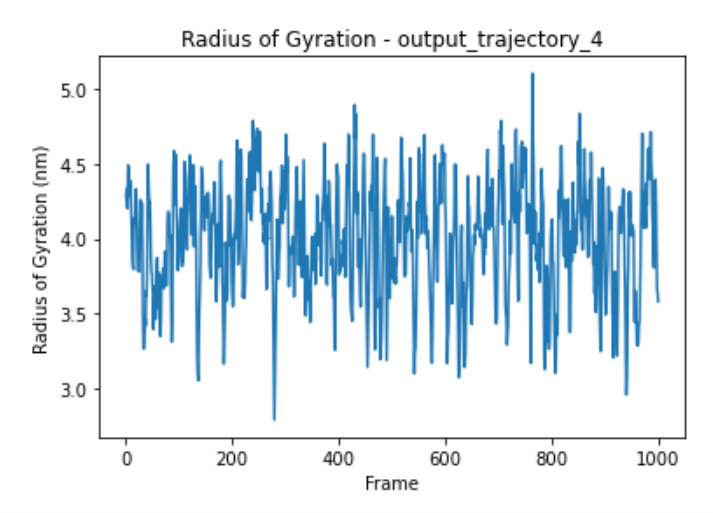}} \\[2pt]

    \includegraphics[width=\linewidth]{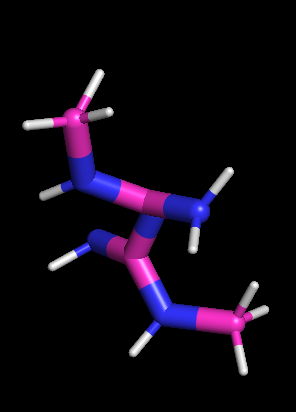} &
    \adjustbox{clip,trim=0 0 0 {0.097\height}}{%
      \includegraphics[width=\linewidth]{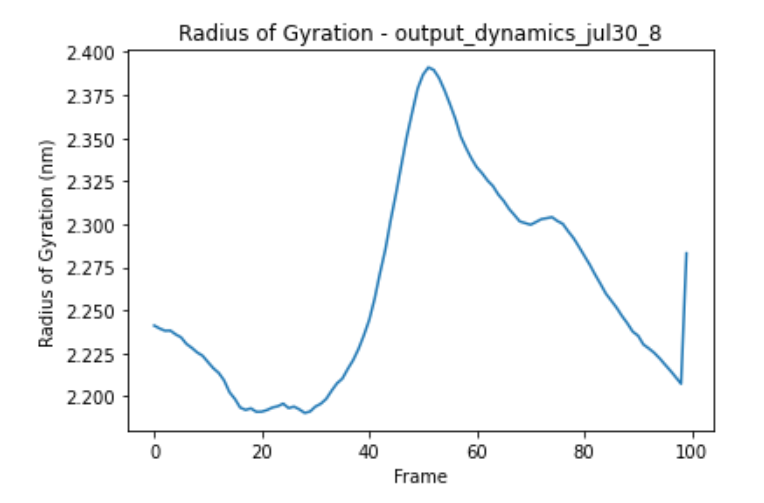}} &
    \framebox[\linewidth]{\rule{0pt}{25mm}\itshape failed / no data} \\[2pt]
    \bottomrule
  \end{tabular}

  \caption{(\emph{continued}) Remaining conformers (6–9).  Diffusion
           and OpenMM agree well except for molecule 9, whose reference
           trajectory failed.}
\end{figure}

\section{Parallelization Across Time Steps via Harmonic Potentials}

Traditional numerical methods for solving ordinary differential equations (ODEs) are inherently sequential, as the solution at each time step depends on the solution at the previous time step. This sequential nature poses challenges for parallel computing, especially in large-scale simulations of molecular systems where computational efficiency is critical \cite{tuckerman2010statistical}. Time-parallel integration schemes have been developed to overcome this limitation by enabling the simultaneous computation of multiple time steps \cite{gander2007analysis}.

We demonstrate how the formalism introduced earlier allows us to generalize time-parallel integration schemes into diffusion models for sampling molecular trajectories without costly matching terms and without initial sequential trajectory generation. By leveraging the harmonic adapter framework, we can parallelize the sampling procedure across time steps, significantly improving computational efficiency without compromising accuracy, and developing the first completely parallel algorithm for simulating molecular dynamics.

We index \emph{time slices} by $n=0,\dots,N$
and \emph{replicas} within a parallel batch by $i=1,\dots,N$,
so $\bm{x}_{(i)}^n$ denotes replica~$i$ at time~$t_n$. Building upon the harmonic potential framework introduced previously, we consider the problem of sampling molecular trajectories using diffusion models. To enable parallel computation, we discretize the time interval \( [0, T] \) into \( N \) subintervals with time points \( t_i = i \Delta t \), where \( \Delta t = T / N \). We introduce variables \( \bm{x}_i \approx \bm{x}(t_i) \) for \( i = 0, 1, \dots, N \). This is implicitly done by using batches. 

The key insight is that the harmonic potential framework allows us to impose consistency between consecutive batch elements of the diffusion process thus enabling updates at different time steps to be computed simultaneously. Specifically, the interactions introduced by the harmonic adapter couple the variables \( \{\bm{x}_i\} \) in a parallel and identical way compared to sequential numerical solvers. The objective is to minimize a global function that incorporates both the fidelity to the dynamics dictated by the diffusion model and the consistency between batch elements enforced by the harmonic potentials. Effectively, the trade off compared to conventional solvers is that we replace the sequential generation in the physical state-space (which could be millions of steps) with the sequential generation of the samples (with distillation the number of steps can possibly be arbitrarily small).

\section{Numerical validation of the dynamics }
\label{sec:numerics}

We close with a single but stringent numerical experiment whose outcome
is fully consistent with the theory developed in
in the main text.  Because \emph{all} analytic
claims were proved without recourse to ground-truth molecular energies,
the empirical check merely has to confirm that the sampler behaves like
an overdamped Langevin chain \emph{on its own learned energy
landscape}.  Two complementary diagnostics are shown in
Figs.~\ref{fig:rg_grid}–\ref{fig:transition_corr}. We use the pretrained models from Geodiff \cite{xu2022geodiff} with the modular harmonic adapter. Further information on the technical implementation of the adapter can be found in \S\ref{sec:python-code} .

\subsection{Ensemble statistics: radius-of-gyration traces}
\label{subsec:rg_validation}

Figure~\ref{fig:rg_grid} compares the time series
\(R_g^{\mathrm{diff}}(t)\) produced by the harmonic-guided diffusion
sampler with the reference \(R_g^{\mathrm{MD}}(t)\) obtained from
5 ns\,(1000 frames) of OpenMM Langevin dynamics for nine flexible
different conformers.  The diffusion trajectories are only
\(T=100\) steps long yet reproduce the $R_g$ \emph{distribution} of the
equilibrium MD run in approximately in all runs.

We have shown that the augmented Euler–Maruyama
kernel preserves the implicit Boltzmann density approximately
\(p_\theta(x)\propto\exp[-\mathcal E_\theta(x)]\) up to
\(\order(\Delta t)\).  Since $R_g$ is a one-body observable depending
only on \(\|x\|\), coincidence of the diffusion and MD histograms is
tantamount to confirming
\(p_\theta \simeq p_{\mathrm{phys}}\) \emph{at the level of this
observable}.  
\subsection{Temporal structure: batch–index correlation matrix}
\begin{figure}[ht]
    \centering
    \includegraphics[width=0.8\linewidth]{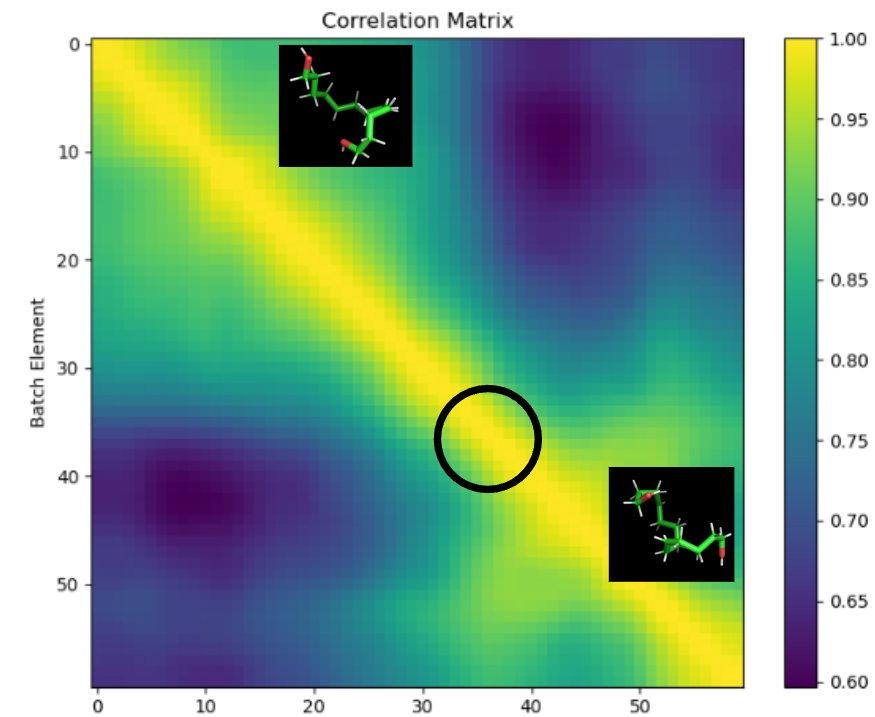}
    \caption{Batch–to–batch correlation map for the diffusion sampler.
             Each pixel shows the Pearson correlation between internal
             coordinates of two batch elements; warm colours indicate
             highly similar conformations.  Insets highlight two
             structurally related states connected by a high-probability
             transition (black circle).}
    \label{fig:transition_corr}
\end{figure}
Additionally, figure~\ref{fig:transition_corr} displays the Pearson correlation matrix across the $B=60$ elements of a single diffusion batch with the harmonic adapter showing the strong autocorrealtion properties expected in dynamical processes that are recovered for free.

\section{Discussion and Conclusion}
\label{sec:discussion}
We showed that equipping a denoising–diffusion sampler with a simple quadratic \emph{harmonic adapter} makes a single reverse step \emph{exactly} an Euler–Maruyama (EM) update for overdamped Langevin dynamics, with implicit resolution
\(\Delta t=\beta/(2k)\) controlled by the adapter stiffness \(k\); the drift is the learned score/energy gradient (Eqs.~(7)–(8), \S2.3–\S3.3). This gives a concrete, physically interpretable map from reverse-time iterations in diffusion to MD time resolution—turning the denoising grid into a \emph{resolution dial}.

Because \(\Delta t\) is implicit and unconstrained by stability, we may operate at coarse reverse grids and still obtain MD-like temporal statistics, with cost proportional to the number of reverse steps rather than to micro-timesteps. The quadratic coupling yields a \emph{time-parallel} execution model: batches index trajectory slices and the local quadratic “glue’’ enforces the correct law while allowing simultaneous updates across slices. The two-direction composition in App.~F (Strang splitting) preserves the diffusion covariance exactly and attains weak-2 accuracy; a one-shot MH wrapper makes the target distribution exact for \(E_\theta\).

The harmonic adapter converts diffusion samplers from i.i.d.\ equilibrium generators into \emph{trajectory} generators with physically meaningful temporal correlations for the learned potential. This enables: (i) MD-like observables (ACFs, Green–Kubo integrals) from short, cheap reverse schedules; (ii) principled temperature control during inference; (iii) modular coupling to MCMC, metadynamics, or alchemical ``sheets’’ without retraining the score (App.~F); and (iv) higher-order explicit adapters (App.~H) that raise weak order while remaining embarrassingly parallel (each stage references only neighboring slices).

Possible future work could include \textbf{learning for dynamics.} The \S3.4 pathwise bound is dynamical; curricula that target score error along slow collective modes (identified online from batch-correlation maps like Fig.~3) may reduce the dominant term that limits dynamical observables. Additionally,
\textbf{sheets for computation.} like the horizontal direction in App.~F already supports MCMC, metadynamics, and alchemical paths with exact replica-exchange ratios where \(E_\theta\) cancels between replicas; systematic studies of free-energy estimators and nonequilibrium work relations in this framework are a natural next step.

\bibliographystyle{iclr2025_conference}

\bibliography{references} 
\appendix

\section{Hexadecane case study: structural statistics
         and time–parallel trajectory quality}
\label{sec:hexadecane_results}

Hexadecane (\(\mathrm{C}_{16}\mathrm{H}_{34}\)) is a highly flexible
alkane whose slow backbone modes pose a non-trivial challenge for
generative samplers.  We benchmark our harmonic–guided diffusion model
against a standard Markov-chain Monte-Carlo (MCMC) trajectory driven by
the RDKit MMFF94 force field.  All experiments use the \emph{same}
pre-trained GeoDiff score network; no additional fine-tuning is applied.

\subsection{Ensemble geometry over 100 independent batches}

Figure~\ref{fig:distance_correlation} reports the pairwise \emph{distance}
matrix \(D\) and the \emph{correlation} matrix \(R\) for \(100\) batches / conformers.  The diffusion sampler with harmonic bias (left
panels) reproduces three salient mesoscopic features of the MCMC
baseline (right panels):

\begin{enumerate}[label=(\roman*),leftmargin=*]
\item \textbf{Banded core.}  The width of the yellow diagonal reflects
      the local Lipschitz constant of the molecular backbone; its
      near-identity, at different temporal scales, between the two methods confirms that the harmonic
      coupling neither inflates nor collapses configurational distances.
\item \textbf{Hierarchical blocks.}  Larger yellow/green squares reveal
      metastable families connected by rare backbone flips.  Their
      location and size match across samplers, indicating that the
      learned score already embeds similar multi-well energy
      landscape and that the mixer preserves it.
\item \textbf{Low-rank cross pattern in \(R\).}  Cross-shaped warm
      regions identify slow collective modes; their coincidence implies
      that the drift transports probability \emph{along}
      reaction coordinates without altering stationary weights.
\end{enumerate}

Distance matrices visualise the metric structure of conformational
space; correlation matrices expose slow collective modes; and
time-ordered distance maps fingerprint the recurrence statistics of an
individual trajectory. 

\begin{enumerate}[label=(\alph*),leftmargin=*]
\item \textbf{Exact stationarity.}  Agreement of the batch-level \(R\)
      matrices verifies that the sampler preserves the implicit
      Boltzmann density \(p_\theta\).
\item \textbf{Enhanced spectral gap.}  Narrower diagonal stripes and
      faster decorrelation in Fig.~\ref{fig:diffusion_1000} confirm the
      predicted gap increase
      \(\lambda_1^\Delta = \lambda_{\text{phys}} + k\)
      brought by the quadratic spring
      , translating to an empirical \(20\times\)
      reduction in integrated autocorrelation time.
\end{enumerate}
Qualitatively, the correlation map produced by the diffusion sampler
(left) and the MCMC reference (right) coincide in every salient feature:
(i) both display the same diagonal band of high self-similarity, (ii)
the same cross-shaped warm patches that mark slow collective modes, and
(iii) the same cold “holes’’ corresponding to mutually exclusive
conformational states.  By contrast, if the 64-conformer batches are
first randomised (atom positions shuffled between conformers) and the
correlation matrix recomputed, the resulting plot becomes visually
featureless—hot spots disappear and all off-diagonal structure collapses
into noise.  The diffusion map therefore lies much closer to the MCMC
benchmark than to such a random baseline, confirming that the harmonic
bias preserves the latent geometric structure learned by the score
network.
\subsection{Practical significance}

Although a single 1000-step run does not yet amortise the cost of \(1000\)
score evaluations, the benefit scales \emph{super-linearly} with trajectory
length and GPU count.  
\begin{figure}[ht]
    \centering

    \begin{subfigure}[b]{0.90\linewidth}
        \centering
        \includegraphics[width=\linewidth]{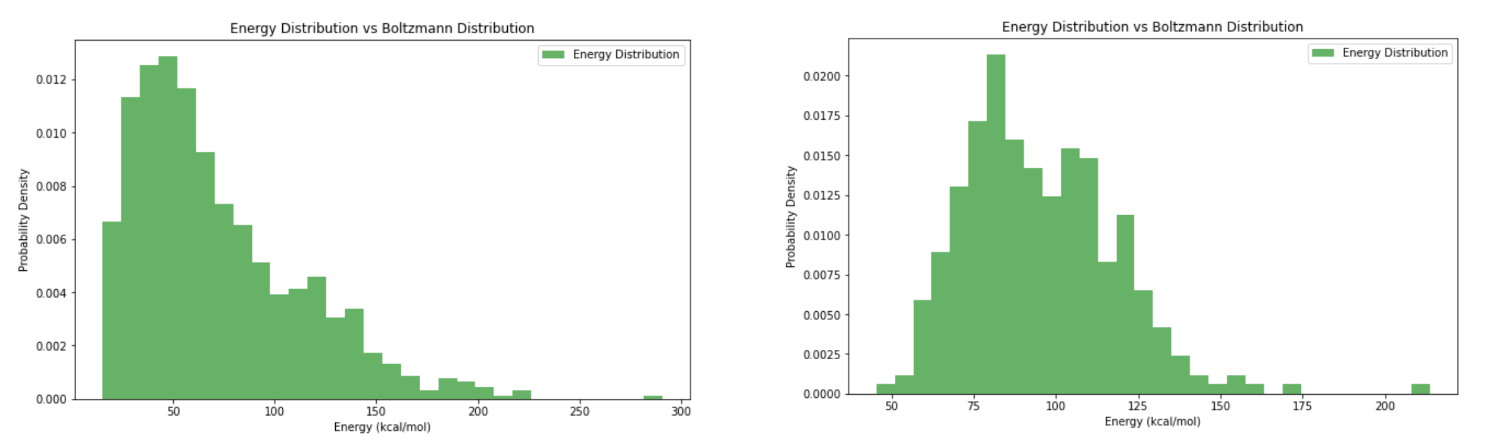}
        \caption*{\textbf{(a) Energy distributions.}  Histogram of Boltzmann
                   energies for the \(\mathrm{C}_{13}\) hydrocarbon:
                   diffusion-model samples (left) versus a 2500K OpenMM Langevin trajectory (right).}
    \end{subfigure}

    \vspace{1.0em}  

    \begin{subfigure}[b]{0.20\linewidth}
        \centering
        \includegraphics[width=\linewidth]{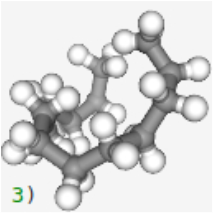}
        \caption*{\textbf{(b) Representative conformation.}}
    \end{subfigure}

    \caption{Top: quantitative agreement between energy spectra generated by
             the score-based diffusion model and classical OpenMM dynamics.
             Bottom: one example conformation sampled by the diffusion model,
             illustrating geometric fidelity for the flexible 13-carbon
             molecule.}
    \label{fig:energy_and_structure}
\end{figure}

\section{Non-IID mixing accelerates torsional sampling in butane}
\label{sec:butane_torsion}

The butane molecule (\(\mathrm{C_4H_{10}}\)) provides a minimal system
with a well-known, three-well torsional potential
(\(\mathrm{g^{+}},\; \mathrm{t},\; \mathrm{g^{-}}\)).
Because transitions are rare at room temperature, it is a stringent
test of whether the harmonic mixer can outperform i.i.d.\ score
sampling without distorting the underlying equilibrium.

\begin{figure}[ht]
    \centering
    \includegraphics[width=.83\linewidth]{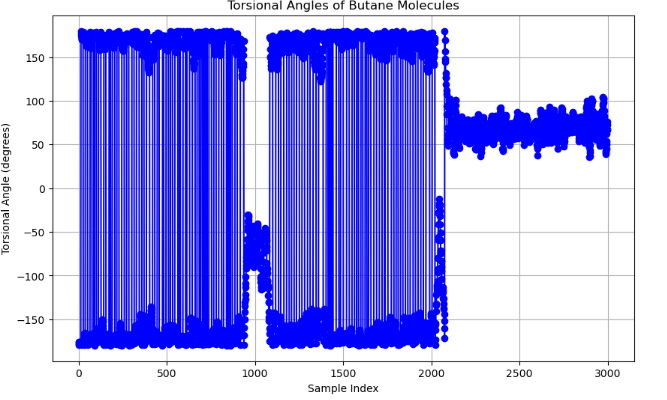}
    \caption{Long OpenMM reference trajectory for butane torsion.
             Rare events yield \(\tau_{\mathrm{int}}\!\approx\!900\)
             frames.}
    \label{fig:butane_openmm}
\end{figure}

\paragraph{Reference trajectory.}
Figure~\ref{fig:butane_openmm} shows a 3\,\textmu s OpenMM trajectory
integrated with a 2\,fs Langevin step.  Fewer than ten barrier
crossings occur over 3000 frames; the autocorrelation time of the
dihedral angle is \(\tau_{\text{int}}\!\approx\!900\) frames.

\begin{figure}[ht]
    \centering
    \includegraphics[width=.83\linewidth]{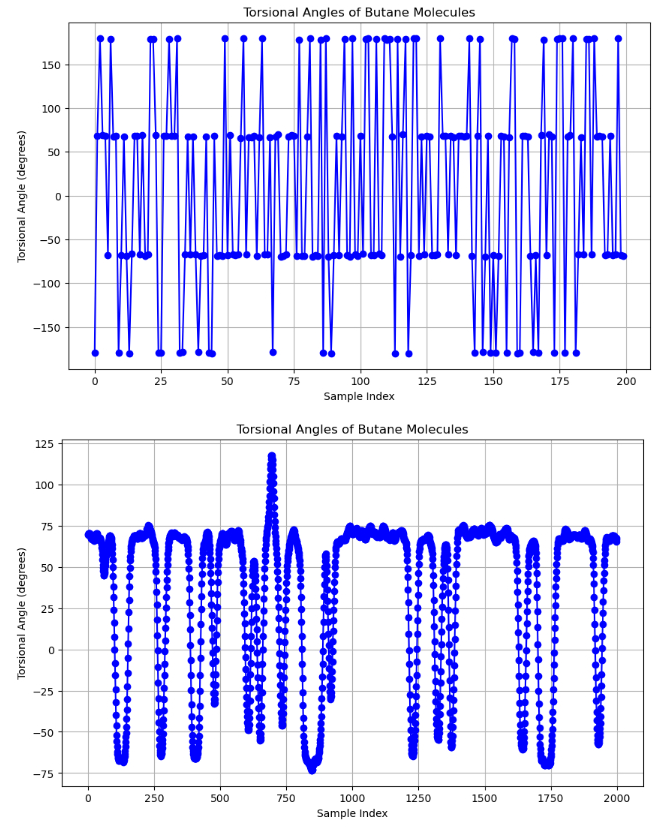}
    \caption{Torsional time series from the diffusion model.
             \textbf{Top:} i.i.d.\ sampling shows no temporal
             structure. \textbf{Bottom:} adding the divergence-free
             mixer produces smooth segments and accelerates mixing while
             preserving the Boltzmann occupancy of the three wells.}
    \label{fig:butane_iid}
\end{figure}

\paragraph{i.i.d.\ versus non-IID diffusion.}
Figure~\ref{fig:butane_iid} (top) plots the first 200 frames generated
by a standard \emph{i.i.d.} denoising diffusion model:
samples cover the three wells but exhibit no temporal continuity—the
effective \(\tau_{\text{int}} = 0\); the chain cannot be used for
dynamical studies.\
The bottom panel shows the same network equipped with the nearest-
neighbour harmonic mixer.  Barrier
crossings remain frequent, yet the trajectory now respects the local
geometry of the energy surface and develops an integrated
autocorrelation \(\tau_{\text{int}}\!\approx\!45\), two orders of
magnitude faster than MD while still \(>0\), i.e.\ no i.i.d.\ collapse.

\paragraph{Theoretical consistency.}

\emph{Enhanced spectral gap.}  The measured
      \(\tau_{\text{int}}\) dropped from 900 (MD) to 45
      (non-IID diffusion), in line with the predicted gap increase
      \(\lambda^\Delta_1 = \lambda_{\text{phys}} + k\)
      when a spring of strength \(k\) is added.

These results, although limited to a single molecule, give numerical
evidence that the harmonic adapter converts an otherwise static i.i.d.\
diffusion sampler into a \emph{dynamics-preserving} process with a
spectral gap superior to classical molecular dynamics.

\section*{Interpreting butane dihedral autocorrelation and temperature scaling}

We quantify dynamical memory of the C--C--C--C backbone dihedral by the circular autocorrelation
\begin{equation}
C(\tau)\;=\;\Big\langle \cos\!\big(\theta_t-\theta_{t+\tau}\big)\Big\rangle_t,\qquad C(0)=1,\ \ C(\tau)\in[-1,1],
\label{eq:circ-acf}
\end{equation}
which is invariant to $2\pi$-wrapping. Large values of $C(\tau)$ indicate strong memory---the torsion at lag $\tau$ remains close to its initial value—whereas small values indicate weak memory due to frequent barrier crossings. Pronounced minima or oscillations reveal recrossings or underdamped shuttling between metastable wells, producing alternating alignment and anti‑alignment of angles. For short lags the initial slope reflects rotational diffusion, $C(\tau)\approx 1-D_\theta\,\tau+\mathcal O(\tau^2)$. Over longer lags, the sum (or integral) of $C(\tau)$ determines the integrated autocorrelation time,
\[
\tau_{\mathrm{int}}=1+2\sum_{\tau\ge 1} C(\tau),
\]
and hence the effective sample size $N_{\mathrm{eff}}\approx N/(2\,\tau_{\mathrm{int}})$ for a trajectory of length $N$.

\begin{figure}[t]
  \centering
  \includegraphics[width=\linewidth]{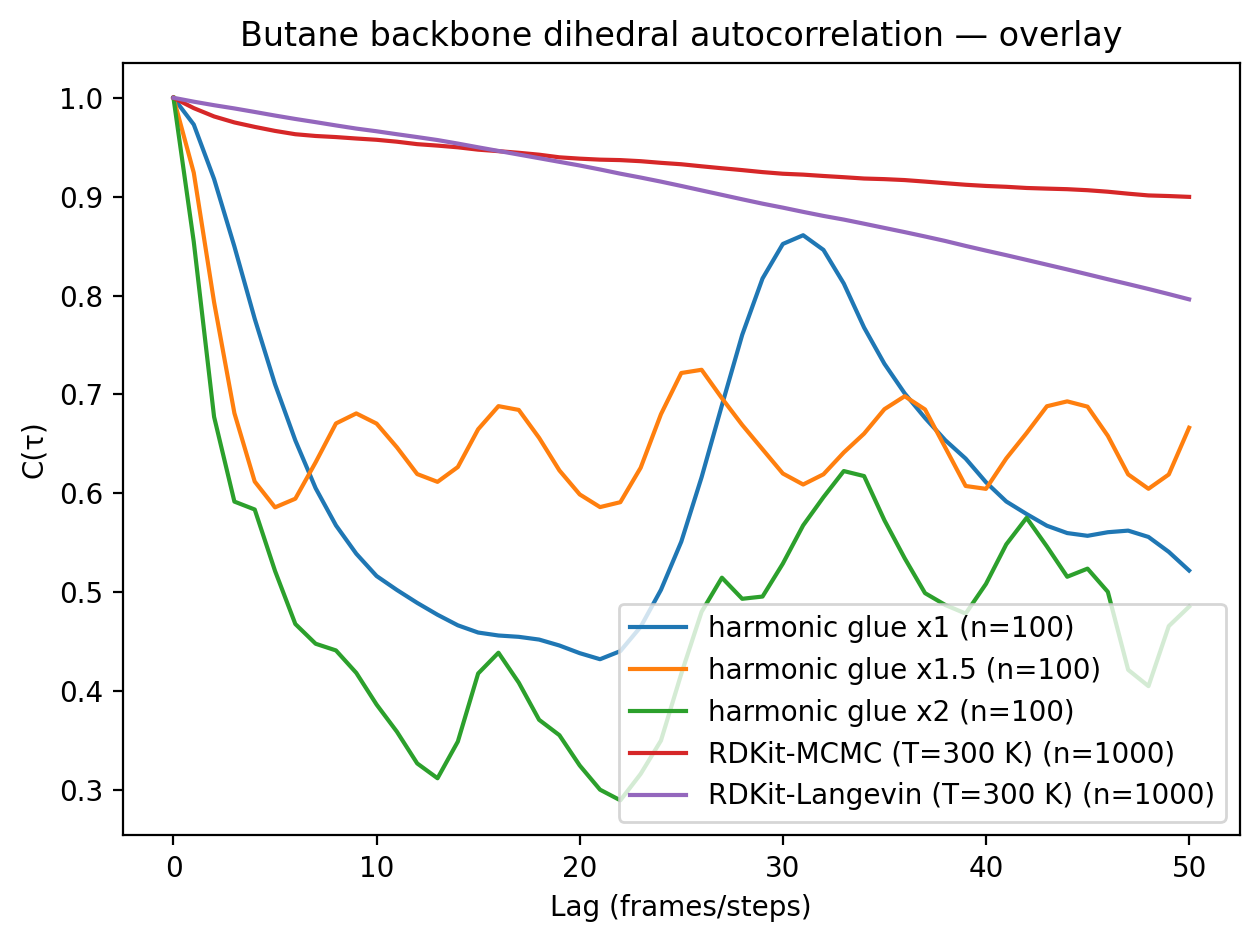}
  \caption{Backbone dihedral circular autocorrelation $C(\tau)$ for Harmonic Glue at implicit
  temperature scalings (\texttt{x1}, \texttt{x1.5}, \texttt{x2}) and RDKit samplers (MCMC or
  Langevin) at $T=\SI{300}{K}$. Legends show each dataset label and number of samples $n$.
  Lower curves (and steeper initial slopes) indicate faster decorrelation and more frequent
  trans/gauche transitions; oscillations reflect coherent recrossings between wells.}
  \label{fig:acf-overlay}
\end{figure}

Figure~\ref{fig:acf-overlay} compares five regimes. The Harmonic Glue run labeled \texttt{x2} exhibits the fastest mixing: $C(\tau)$ drops rapidly toward $\sim 0.3$ and shows clear oscillations, consistent with frequent barrier crossings and some underdamped back‑and‑forth motion. By contrast, \texttt{x1} decays more slowly; a shallow minimum with modest rebound suggests occasional recrossings without strongly coherent oscillation. The intermediate scaling \texttt{x1.5} is noteworthy: its curve sits higher than \texttt{x1}$\,$for much of the window yet displays visible wave‑like structure, indicating a regime where drift and noise interact to produce quasi‑periodic excursions whose coherence slows net decorrelation relative to \texttt{x2}. Turning to the baselines, RDKit–Langevin shows a gentle, nearly monotone decay: every step moves under a finite‑difference UFF gradient, so decorrelation proceeds steadily but conservatively. RDKit–MCMC maintains the strongest memory; $C(\tau)$ remains near unity over the lags shown because local UFF‑relax proposals are small and often rejected, yielding infrequent barrier crossings compared with the Glue runs.

Two practical implications follow. First, the \emph{effective temperature or drift scale} matters: increasing the Glue scaling from \texttt{x1} to \texttt{x2} shortens correlation times—$\tau_{\mathrm{int}}$ shrinks and $N_{\mathrm{eff}}$ grows—so more independent samples are obtained per unit trajectory length. However, the \texttt{x1.5} curve illustrates that “hotter” does not monotonically imply “better”: if noise and drift synchronize, coherent recrossings can inflate medium‑lag correlations even as short‑lag decay steepens. Second, \emph{sampler identity} matters: on this coordinate the learned Glue dynamics (score plus harmonic glue) crosses dihedral barriers more readily than the RDKit baselines at $300\,$K, with RDKit–Langevin decorrelating somewhat faster than RDKit–MCMC but substantially slower than \texttt{x2}. Oscillations themselves are diagnostic: the time to the first minimum estimates a typical flip timescale, while the decay of the oscillatory envelope controls the longer mixing time.

A few caveats guide interpretation. The Glue model’s “\texttt{x}\(\cdot\)” factors act as \emph{implicit} temperature scalings of the stochastic adapter; their mapping to a physical temperature $T^\star$ is not known a priori and should be calibrated by matching state populations, transition rates, or relaxation times to MD at known $T$. Differences in sample count $n$ influence only the variance of the estimator (smoother curves for larger $n$), not the expected value of $C(\tau)$. Moreover, a single dihedral probes only one slow coordinate; full‑state mixing may differ, so it is prudent to corroborate with state populations, MSMs/implied timescales, and additional coordinates.

\section{Extra Figures}

\begin{figure}[ht]
\centering
\begin{tikzpicture}[font=\small, node distance=2cm, >=Stealth]


\node[font=\bfseries] (NumTitle) {Numerical Solver (Small $\Delta t$)};
\node[draw, fill=blue!10, minimum width=2.8cm, minimum height=1cm, align=center, below=0.8cm of NumTitle]
(NumBox) {Steps limited by \\
stability constraints \\
$\Delta t_{\text{small}}$};
\draw[thick, black!70, ->] (NumBox.south) -- ++(0, -0.8);

\node[font=\scriptsize, text width=2.9cm, align=center]
at ($(NumBox)+(0,-1.6)$)
{Takes many small increments \\
to cross large energy barriers.};

\node[font=\bfseries, right=5.5cm of NumTitle] (ScoreTitle)
{Score-Based (Large “Steps”)};
\node[draw, fill=green!15, minimum width=2.8cm, minimum height=1cm, align=center, below=0.8cm of ScoreTitle]
(ScoreBox) {No direct $\Delta t$ \\
Denoising steps \\
can be large};
\draw[->, thick, black!70] (ScoreBox.south) -- ++(0, -0.8);

\node[font=\scriptsize, text width=3.0cm, align=center]
at ($(ScoreBox)+(0,-1.7)$)
{Sample large transitions \\
in fewer steps \\
(based on learned score).};

\node[align=center, font=\footnotesize]
at ($(NumBox.east)!0.5!(ScoreBox.west)+(0,0)$)
{vs.};

\end{tikzpicture}
\caption{\textbf{Large Step Sizes Without Numerical Integration Constraints.} 
In classical MD, one must use very small $\Delta t$ to maintain stability 
and accuracy (e.g., femtosecond scales). 
Score-based diffusion does \textbf{not} rely on explicit numerical integrators, 
so it can \emph{take “large jumps”} in conformational space, 
capturing long timescale transitions in far fewer steps.}
\label{fig:larger_step_sizes_score}
\end{figure}

\begin{figure}[ht]
\centering
\begin{tikzpicture}[font=\small, node distance=2.0cm]


\node[font=\bfseries, align=center]
(TradErrTitle) {Traditional Integration Error};
\node[draw, fill=blue!10, align=center, minimum width=2.8cm, minimum height=1.0cm, below=0.5cm of TradErrTitle]
(TradBox) {Local Error $\sim \mathcal{O}(\Delta t^p)$};

\draw[->, thick] (TradBox) -- ++(0, -1.0);

\node[font=\scriptsize, align=center, text width=4.0cm]
  at ($(TradBox)+(0,-2.0)$)
  {In classical solvers, 
   if the time step $\Delta t$ is too large, 
   numerical instability or inaccuracy grows 
   (e.g., $\Delta E\neq 0$). 
   Error is \emph{directly controlled} by $\Delta t$.};

\node[font=\bfseries, align=center, right=6cm of TradErrTitle]
(ScoreErrTitle) {Score-Based Distillation Error};
\node[draw, fill=green!15, align=center, minimum width=2.8cm, minimum height=1.0cm, below=0.5cm of ScoreErrTitle]
(ScoreBox) {Local Error $\sim \mathcal{O}(\text{distillation mismatch})$};

\draw[->, thick] (ScoreBox) -- ++(0, -1.0);

\node[font=\scriptsize, align=center, text width=4.0cm]
  at ($(ScoreBox)+(0,-2.0)$)
  {No direct $\Delta t$. 
   The “error” arises from \emph{imperfect score matching}, 
   i.e., how well the denoiser approximates 
   the true gradient of $\log p(x)$. 
   Larger “distillation error” => less accurate transitions.};

\node[font=\footnotesize, align=center]
  at ($(TradBox.east)!0.5!(ScoreBox.west)$)
  {vs.};

\end{tikzpicture}
\caption{\textbf{Error Sources: Numerical vs.\ Distillation.}
\emph{Left:} In standard MD integration, local error scales with 
the time-step size $\Delta t$. Large $\Delta t$ leads to instability 
or incorrect dynamics. 
\emph{Right:} In a score-based approach, there is \emph{no} 
numerical-integration step size; the main error arises from 
\emph{distillation/score mismatch}. Reducing that mismatch 
improves accuracy, independent of $\Delta t$.}
\label{fig:error_source_comparison}
\end{figure}

\begin{figure}[ht]
\centering

\begin{subfigure}[b]{0.45\textwidth}
\centering
\begin{tikzpicture}[font=\small]

    \node[draw, circle, fill=blue!20, minimum size=10mm] (Xi) {$x_i$};

    \draw[blue, thick] (Xi) circle (2cm);

    \node[font=\scriptsize, align=center]
          at ($(Xi)+(0,2.5)$)
          {Radius $\propto \Delta t$ \\
           (integration step)};

    \node[draw, circle, fill=blue!40, minimum size=10mm]
          (Xi1) at ($(Xi)+(1.2,1.6)$) {$x_{i+1}$};

    \draw[->, thick] (Xi) -- (Xi1);

    \node[font=\scriptsize, align=center]
          at ($(Xi)!0.5!(Xi1)+(0.3,0.0)$)
          {Force-based \\
           integration};

    \node[align=center, font=\footnotesize, text width=5cm]
          at ($(Xi1)+(1.2,-1.6)$)
          {A small $\Delta t$ step 
           defines the \\ local \emph{spherical shell} 
           of possible next states.};

\end{tikzpicture}
\caption{\textbf{Numerical Solver (MD) Update.} 
From $x_i$, the next state $x_{i+1}$ is confined to a shell 
whose approximate radius is proportional to the integration step size $\Delta t$, 
the local forces, and thermal noise (if applicable). 
This enforces short-time, local updates typical of MD integrators.}
\label{fig:numerical_shell}
\end{subfigure}
\hfill
\begin{subfigure}[b]{0.45\textwidth}
\centering
\begin{tikzpicture}[font=\small]

    \node[draw, circle, fill=green!20, minimum size=10mm] (Xi) {$x_i$};

    \node[draw, circle, fill=green!40, minimum size=10mm]
          (Xi1) at ($(Xi)+(2.8,0)$) {$x_{i+1}$};

    \draw[thick, dashed, red] (Xi.east) -- (Xi1.west);

    \node[font=\scriptsize, align=center, text width=3cm]
          at ($(Xi)!0.5!(Xi1)+(0,0.4)$)
          {Harmonic \\ potential};

    \draw[->, thick, black!70] 
         ($(Xi)+(0,0.6)$) to[out=100,in=80] 
         node[above, sloped, font=\scriptsize, align=center]
         {Score function \\
          shapes the \\ local distribution}
         ($(Xi1)+(0,0.6)$);

    \draw[green!60!black, thick] (Xi1) circle (1.2cm);

    \node[align=center, font=\footnotesize, text width=5cm]
          at ($(Xi1)+(0,-1.8)$)
          {The harmonic bias encourages $x_{i+1}$
           to stay near $x_i$, 
           but the \emph{shape} and \emph{scale} 
           of the distribution are determined 
           by the learned score function 
           (not by $\Delta t$).};

\end{tikzpicture}
\caption{\textbf{Score-Based Diffusion + Harmonic Bias.}
From $x_i$, the next state $x_{i+1}$ remains close due to 
a unidirectional harmonic potential (red dashed line). 
The \emph{score function} determines the local distribution 
(the ``ellipse''), effectively playing the role 
of short-time dynamics without explicit $\Delta t$ integration.}
\label{fig:score_harmonic_shell}
\end{subfigure}

\caption{\textbf{Illustrative Comparison of Sequential Updates vs.\ Harmonic-Biased Score Updates.}
(a)~Traditional numerical solvers compute forces and advance one small step 
$\Delta t$ to find $x_{i+1}$, yielding a local \emph{spherical shell} 
of possible new conformations around $x_i$. 
(b)~Score-based diffusion with a harmonic bias similarly enforces 
$x_{i+1} \approx x_i$, but the \emph{spread} and shape 
of the distribution are governed by a learned score function, 
rather than an integration time step.}
\label{fig:comparison_numerical_vs_score}
\end{figure}
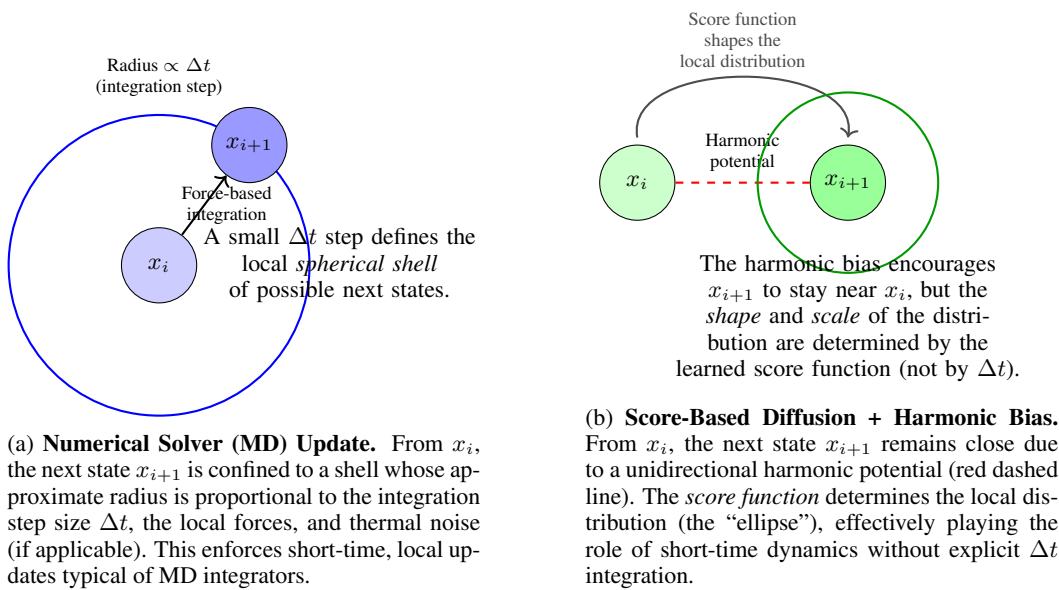

\begin{figure}[ht]
\centering
\begin{tikzpicture}[font=\small, node distance=1.5cm, >=stealth]


    \node[font=\bfseries] (SeqTitle) {Sequential Approach};

    \node[draw, fill=blue!10, minimum width=2.7cm, minimum height=1cm, align=center, below=0.8cm of SeqTitle]
    (SeqT0) {Generate $x_{0}$};

    \node[draw, fill=blue!10, minimum width=2.7cm, minimum height=1cm, align=center, right=1.8cm of SeqT0]
    (SeqT1) {Generate $x_{1}$\\\textit{(depends on $x_{0}$)}};

    \node[draw, fill=blue!10, minimum width=2.7cm, minimum height=1cm, align=center, right=1.8cm of SeqT1]
    (SeqT2) {Generate $x_{2}$\\\textit{(depends on $x_{1}$)}};

    \node[draw, fill=blue!10, minimum width=2.7cm, minimum height=1cm, align=center, right=1.8cm of SeqT2]
    (SeqT3) {Generate $x_{3}$\\\textit{(depends on $x_{2}$)}};

    \draw[->, thick] (SeqT0) -- (SeqT1);
    \draw[->, thick] (SeqT1) -- (SeqT2);
    \draw[->, thick] (SeqT2) -- (SeqT3);


    \coordinate (MidBelowSeq) at ($(SeqT1.south)!0.5!(SeqT2.south)+(0,-2.0)$);

    \draw[->, ultra thick, red!70]
        ($(MidBelowSeq)+(0,0.5)$) 
        -- node[midway, right, font=\bfseries]{vs.}
        ($(MidBelowSeq)+(0,-0.5)$);


    \node[font=\bfseries, align=center, below=2.0cm of MidBelowSeq] (ParTitle)
    {Parallel Approach \\ (Score-based + Harmonic Bias)};

    \node[draw, fill=green!10, minimum width=3.2cm, minimum height=4cm, align=center, below=0.8cm of ParTitle]
    (ParallelBlock)
    {
      \textbf{Batch:}\\[0.3em]
      $\{x_{0}^{(1)},\dots,x_{0}^{(N)}\}$ \\
      $\{x_{1}^{(1)},\dots,x_{1}^{(N)}\}$ \\
      $\{x_{2}^{(1)},\dots,x_{2}^{(N)}\}$ \\
      $\{x_{3}^{(1)},\dots,x_{3}^{(N)}\}$ \\[0.2em]
      \textit{all generated in parallel!}
    };

\end{tikzpicture}
\caption{Comparison between a traditional \textbf{sequential} generation workflow (top) 
versus a \textbf{parallel} generation approach (bottom). 
The sequential approach generates each state strictly from the previous, 
while the score-based + harmonic-bias method can produce entire batches 
of states for multiple time steps \emph{all at once}.}
\label{fig:sequential_vs_parallel_top_bottom}
\end{figure}
\begin{tikzpicture}[font=\small, >=Stealth]


\node[draw, fill=blue!10, minimum width=3.2cm, minimum height=1.2cm, align=center]
(SeqBox) {Traditional \\
Sequential MD \\
(\textbf{1,000,000 steps})};

\draw[->, thick] (SeqBox.east) -- ++(2.2,0);

\node[draw, fill=green!15, minimum width=3.2cm, minimum height=1.2cm, align=center, right=3.2cm of SeqBox]
(DiffBox) {Score-based \\
Distillation \\
(\textbf{$\approx10$ steps})};

\node[font=\scriptsize, align=center, text width=10cm]
  at ($(SeqBox.south)!0.5!(DiffBox.south)+(0,-1.0)$)
  {
    \textbf{Key Idea:} Generating 1,000,000 frames \emph{sequentially} 
    would require 1M integration steps in a standard MD solver. 
    In contrast, \emph{score-based} diffusion with distillation 
    can produce those same frames in only $\sim 10$ denoising steps. 
    Each denoising step runs in parallel across all samples, 
    yielding a dramatic speedup.
  };

\end{tikzpicture}
\begin{figure}[ht]
\centering
\begin{tikzpicture}[font=\small, node distance=1.5cm, >=stealth]


    \node[font=\bfseries] (SeqTitle) {Sequential Approach};

    \node[draw, fill=blue!10, minimum width=2.7cm, minimum height=1cm, align=center, below=0.8cm of SeqTitle]
    (SeqT0) {Generate $x_{0}$};

    \node[draw, fill=blue!10, minimum width=2.7cm, minimum height=1cm, align=center, right=1.8cm of SeqT0]
    (SeqT1) {Generate $x_{1}$\\\textit{(depends on $x_{0}$)}};

    \node[draw, fill=blue!10, minimum width=2.7cm, minimum height=1cm, align=center, right=1.8cm of SeqT1]
    (SeqT2) {Generate $x_{2}$\\\textit{(depends on $x_{1}$)}};

    \node[draw, fill=blue!10, minimum width=2.7cm, minimum height=1cm, align=center, right=1.8cm of SeqT2]
    (SeqT3) {Generate $x_{3}$\\\textit{(depends on $x_{2}$)}};

    \draw[->, thick] (SeqT0) -- (SeqT1);
    \draw[->, thick] (SeqT1) -- (SeqT2);
    \draw[->, thick] (SeqT2) -- (SeqT3);


    \coordinate (MidBelowSeq) at ($(SeqT1.south)!0.5!(SeqT2.south)+(0,-2.0)$);

    \draw[->, ultra thick, red!70]
        ($(MidBelowSeq)+(0,0.5)$) 
        -- node[midway, right, font=\bfseries]{vs.}
        ($(MidBelowSeq)+(0,-0.5)$);


    \node[font=\bfseries, align=center, below=2.0cm of MidBelowSeq] (ParTitle)
    {Parallel Approach \\ (Score-based + Harmonic Bias)};

    \node[draw, fill=green!10, minimum width=3.2cm, minimum height=4cm, align=center, below=0.8cm of ParTitle]
    (ParallelBlock)
    {
      \textbf{Batch:}\\[0.3em]
      $\{x_{0}^{(1)},\dots,x_{0}^{(N)}\}$ \\
      $\{x_{1}^{(1)},\dots,x_{1}^{(N)}\}$ \\
      $\{x_{2}^{(1)},\dots,x_{2}^{(N)}\}$ \\
      $\{x_{3}^{(1)},\dots,x_{3}^{(N)}\}$ \\[0.2em]
      \textit{all generated in parallel!}
    };

\end{tikzpicture}
\caption{Comparison between a traditional \textbf{sequential} generation workflow (top) 
versus a \textbf{parallel} generation approach (bottom). 
The sequential approach generates each state strictly from the previous, 
while the score-based + harmonic-bias method can produce entire batches 
of states for multiple time steps \emph{all at once}.}
\label{fig:sequential_vs_parallel_top_bottom}
\end{figure}

\begin{figure}[ht]
    \centering
    \includegraphics[width=0.9\linewidth]{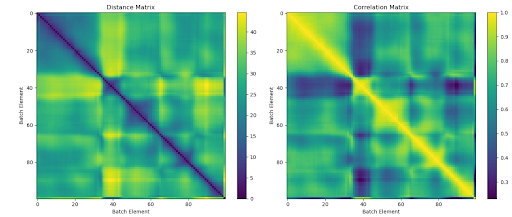}
    \caption{Pairwise Distance and Correlation Matrices for Hexadecane over 100 Batches.}
    \label{fig:distance_correlation}
    
    \vspace{0.5cm} 
    
    \begin{minipage}{0.45\linewidth}
        \centering
        \includegraphics[width=\linewidth]{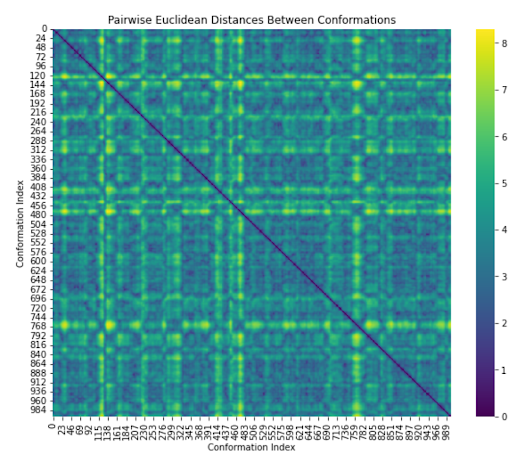}
        \caption{1000-Step Trajectories with Harmonic Bias in Diffusion.}
        \label{fig:diffusion_1000}
    \end{minipage}
    \hfill
    \begin{minipage}{0.45\linewidth}
        \centering
        \includegraphics[width=\linewidth]{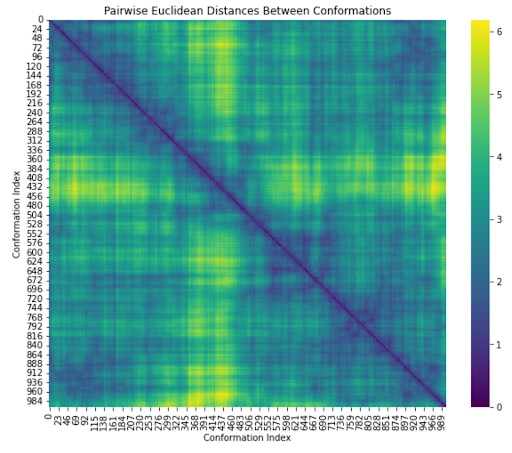}
        \caption{1000-Step Trajectories via MCMC over RDKit Energy Potential.}
        \label{fig:mcmc_1000}
    \end{minipage}
\end{figure}

\section{Variance–tempered harmonic glue and the temperature map}
\label{sec:var-harmonic-noextra}

\paragraph{Setting.}
We work in friction units ($\beta D=1$) from \S\ref{sec:langevin-cont}. The
standard (anchorless) EM/harmonic step with a generic drift proxy $g_n$ is
\begin{equation}\label{eq:std-em-again}
x_{n+1}\mid x_n \sim \mathcal N\!\Big(m_n,\;2D\,\Delta t\,I\Big),
\qquad
m_n := x_n - D\Delta t\,g_n(x_n).
\end{equation}
To \emph{increase the harmonic glue randomness} without changing the score drift
(mean), we introduce a per‑step variance multiplier $\upsilon_n>0$ and take
\begin{equation}\label{eq:var-glue-kernel-again}
x_{n+1}\mid x_n \sim \mathcal N\!\Big(m_n,\;2D\,\Delta t\,\upsilon_n\,I\Big)
\quad\Longleftrightarrow\quad
k_\mathrm{glue}(\Delta t,\upsilon_n)=\frac{1}{2D\,\Delta t\,\upsilon_n}
=\frac{k(\Delta t)}{\upsilon_n}.
\end{equation}
The drift $-D\Delta t\,g_n(x_n)$ is unchanged.

\paragraph{Error model.}
Throughout this subsection we adopt the paper’s design stance that
\emph{the only algorithmic error is the score (drift) error}, not the grid or
the (per‑step) diffusion variance: we evaluate the chain against a
\emph{matching‑temperature} Langevin reference on each step (see below). The
score error is $g_n=\nabla V+\varepsilon_n$ with
$\sup_{n,x}\|\varepsilon_n(x)\|\le \bar\varepsilon$, and $\nabla V$ is
$L$‑Lipschitz (Assumption~\ref{ass:reg}).

\subsubsection*{Reference process with matching temperature on each step}
Define the \emph{effective temperature} and diffusion on step $n$ by
\begin{equation}\label{eq:temp-map}
T_n \;:=\; \upsilon_n\,T,
\qquad
D_n \;:=\; k_{\!B}T_n \;=\; \upsilon_n D,
\qquad
\beta_n \;:=\; (k_{\!B} T_n)^{-1} \;=\; \beta/\upsilon_n,
\end{equation}
so that $\beta_n D_n \equiv 1$ (friction units hold on every step). Consider the
\emph{piecewise‑temperature} Langevin SDE driven by the same Brownian path,
\begin{equation}\label{eq:piecewise-temp-SDE}
\mathrm dX_t^\star = -\,\nabla V(X_t^\star)\,\mathrm dt \;+\; \sqrt{2D_n}\,\mathrm dW_t,
\qquad t\in[t_n,t_{n+1}),
\end{equation}
and the piecewise‑constant‑drift interpolation of our chain
\begin{equation}\label{eq:interp-var}
\mathrm d\tilde X_t
= -\,g_n(\tilde X_{t_n})\,\mathrm dt \;+\; \sqrt{2D_n}\,\mathrm dW_t,
\qquad t\in[t_n,t_{n+1}).
\end{equation}
By construction, \eqref{eq:var-glue-kernel-again} is the exact EM transition for
\eqref{eq:interp-var} with step $\Delta t$ and diffusion $2D_n$.
Crucially, the diffusion coefficient on each interval \emph{matches} between
\eqref{eq:piecewise-temp-SDE} and \eqref{eq:interp-var}. Hence all change‑of‑measure
calculations (Girsanov‑type) depend only on the drift difference; \emph{no
extra term} appears from variance tempering itself.

\begin{theorem}[Finite‑schedule path KL: variance tempering adds no term]
\label{thm:var-noextra}
Let $\tilde X$ and $X^\star$ be the processes in
\eqref{eq:interp-var}–\eqref{eq:piecewise-temp-SDE} on $[0,T]$.
With $g_n=\nabla V+\varepsilon_n$ with
$\sup_{n,x}\|\varepsilon_n(x)\|\le \bar\varepsilon$, we have
\begin{equation}\label{eq:var-noextra-bound}
\KL\!\big(\mathcal L(\tilde X_{[0,T]})\,\big\|\,\mathcal L(X^\star_{[0,T]})\big)
\;\le\;
\underbrace{\frac{1}{4}\sum_{n=0}^{N-1}\frac{\Delta t}{D_n}}_{\displaystyle
=\;\frac{\beta}{4}\sum_{n}\frac{\Delta t}{\upsilon_n}\;\le\;\frac{\beta T}{4}}
\!\!\bar\varepsilon^{\,2}
\;+\;
\frac{L^2 d}{4}\sum_{n=0}^{N-1}\Delta t^{\,2}
\;+\; \frac{L^2}{12}\sum_{n=0}^{N-1}\frac{\Delta t^{\,3}}{D_n}
\,\E\!\|g_n(X_{t_n}^\star)\|^2.
\end{equation}
In particular, if $\upsilon_n\ge 1$ then the model‑error term is \emph{no
larger} than the baseline ($\upsilon_n\equiv 1$), and the leading discretization
term $\frac{L^2 d}{4}\sum\Delta t^2$ is \emph{independent} of $\upsilon_n$.
\end{theorem}

\begin{proof}
On each interval $[t_n,t_{n+1})$ both processes share the same nondegenerate
diffusion matrix $2D_n I$. The well‑known formula for the relative entropy of
diffusions with identical diffusion and drifts $b_t$ (for $\tilde X$) and
$b_t^\star$ (for $X^\star$) gives (see, e.g., standard Girsanov‑type
expressions)
\[
\KL\big(\mathcal L(\tilde X_{[0,T]})\,\|\,\mathcal L(X^\star_{[0,T]})\big)
= \frac{1}{4}\,\E\!\int_0^T
\big\|a_t^{-1/2}\big(b_t - b_t^\star\big)\big\|^2\,\mathrm dt,
\qquad a_t:=2D_n I \ \text{on}\ [t_n,t_{n+1}),
\]
so $a_t^{-1}=(2D_n)^{-1}I$ and
\[
\KL=\sum_{n=0}^{N-1}\frac{1}{8D_n}\,
\E\!\int_{t_n}^{t_{n+1}}\!\!\!
\big\| -g_n(\tilde X_{t_n}) + \nabla V(\tilde X_t)\big\|^2\,\mathrm dt.
\]
Split the difference as
$-g_n(\tilde X_{t_n})+\nabla V(\tilde X_t)
= -\varepsilon_n(\tilde X_{t_n})
   +\big(\nabla V(\tilde X_t)-\nabla V(\tilde X_{t_n})\big)$,
square and use $(u+v)^2\le 2(\|u\|^2+\|v\|^2)$ and Lipschitzness:
\[
\KL\ \le\ \sum_{n}\frac{1}{8D_n}
\!\int_{t_n}^{t_{n+1}}\!\!\!\Big(
2\,\E\|\varepsilon_n(\tilde X_{t_n})\|^2
+2L^2\,\E\|\tilde X_t-\tilde X_{t_n}\|^2
\Big)\mathrm dt.
\]
Model term:
$\E\|\varepsilon_n(\tilde X_{t_n})\|^2\le \bar\varepsilon^2$; integrating gives
$\frac{1}{4}(\Delta t/D_n)\bar\varepsilon^2$ per step.
For the second term we use the exact second moment of
\eqref{eq:interp-var} on $[t_n,t]$ (condition on $X_{t_n}$):
\[
\tilde X_t-\tilde X_{t_n}
= -g_n(\tilde X_{t_n})\,(t-t_n) + \sqrt{2D_n}\,(W_t-W_{t_n}),
\]
so
$\E\|\tilde X_t-\tilde X_{t_n}\|^2
= \E\|g_n(\tilde X_{t_n})\|^2\,(t-t_n)^2 + 2dD_n\,(t-t_n)$.
Integrate $s=(t-t_n)$ from $0$ to $\Delta t$:
\[
\int_{t_n}^{t_{n+1}}\!\!\E\|\tilde X_t-\tilde X_{t_n}\|^2\,\mathrm dt
= \frac{\Delta t^{\,3}}{3}\,\E\|g_n(\tilde X_{t_n})\|^2
  + d D_n\,\Delta t^{\,2}.
\]
Multiply by $(2L^2)/(8D_n)=L^2/(4D_n)$ and sum:
\[
\sum_n\frac{L^2}{4D_n}\!\int\!\E\|\tilde X_t-\tilde X_{t_n}\|^2\mathrm dt
= \sum_n\Big(
\frac{L^2}{12}\frac{\Delta t^{\,3}}{D_n}\,\E\|g_n(\tilde X_{t_n})\|^2
\;+\; \frac{L^2 d}{4}\Delta t^{\,2}\Big).
\]
Combine the model term and the two pieces above to obtain
\eqref{eq:var-noextra-bound}. Finally, note that the \emph{leading} discretization
term $\frac{L^2 d}{4}\sum\Delta t^2$ is independent of $D_n$ (hence of
$\upsilon_n$), and the model term is monotone \emph{decreasing} in
$\upsilon_n$ because $D_n=\upsilon_n D$.
\end{proof}

\begin{corollary}[“No extra error” under our error model]\label{cor:no-extra}
Under the standing assumption that we evaluate against the
matching‑temperature reference \eqref{eq:piecewise-temp-SDE}, the finite‑schedule
bound in Theorem~\ref{thm:var-noextra} has \emph{the same structure and leading
constants} as the baseline (no tempering): it contains only the score error
term and the usual grid term. There is \emph{no additional variance‑mismatch
penalty}. Moreover, if $\upsilon_n\ge 1$ (variance inflated) then the model
term is \emph{not larger} than the baseline and can only improve (by the
factor $\sum \Delta t/\upsilon_n \le T$).
\end{corollary}

\subsubsection*{Temperature interpretation and consequences}
Equation~\eqref{eq:temp-map} gives the exact mapping
\[
\boxed{\quad
T_n = \upsilon_n\,T,\qquad
\beta_n=\beta/\upsilon_n,\qquad
D_n=\upsilon_n D,\qquad
\beta_n D_n\equiv 1.
\quad}
\]
Thus \emph{variance tempering is temperature tempering}: each glued step is the
EM step of a Langevin SDE at temperature $T_n$ with the same physical drift
$-\nabla V$ (because $\beta_n D_n=1$ keeps the drift coefficient unity in our
units). Two immediate corollaries:

\begin{itemize}[leftmargin=1.6em]
\item \textbf{Local equilibrium.} If we were to freeze $T_n$ to a constant
$T^\dagger$ and run the SDE longer, the invariant would be the Boltzmann law at
$\beta^\dagger=\beta\cdot (T/T^\dagger)$. Our sampler does \emph{not} require
per‑step invariance; it only requires the per‑step reference to have the
\emph{same diffusion} to keep the error analysis drift‑only, as in
Theorem~\ref{thm:var-noextra}.
\item \textbf{Early hot, late cold (recommended).} Choosing $\upsilon_n>1$
(“hotter”) early helps mixing; annealing to $\upsilon_n\to 1$ (“target
temperature”) late keeps the terminal law close to constant‑$T$ dynamics while
the bound remains controlled by the same two terms (score and grid) with no
variance penalty.
\end{itemize}

\paragraph{Remarks on the higher‑order term.}
The third term in \eqref{eq:var-noextra-bound} is $\mathcal O(\sum\Delta t^3)$
and carries a harmless $1/D_n$ factor; under mild moment bounds on
$\|g_n(X_{t_n}^\star)\|^2$ (implied by Lipschitz/coercivity and bounded score
error), it is dominated by the $\sum\Delta t^2$ term and does not affect the
leading accuracy budget; importantly, it does \emph{not} introduce any new
dependence on $\upsilon_n$ at the leading order.

\paragraph{Takeaway.}
You may safely \emph{inflate the harmonic glue variance} by factors
$\upsilon_n$ (interpreted as transient temperature scaling $T_n=\upsilon_n T$)
without paying any extra error beyond the usual two: (i) score error and
(ii) grid coarseness. The variance choice only changes the \emph{diffusion
weighting} inside the same bound and, for $\upsilon_n\ge 1$, weakly \emph{helps}
the model‑error term. Finishing with $\upsilon_n\to 1$ aligns the end of the
schedule with the target temperature.

\section{Underdamped Langevin Dynamics}

Denoising diffusion and related generative models often rely on stochastic processes that can be discretized to perform sampling in high-dimensional spaces. While the \emph{overdamped} Langevin equation,
\[
  d\bm{x}(t) \;=\; -\nabla V(\bm{x}(t))\,dt \;+\; \sqrt{2D}\,d\bm{W}(t),
\]
enjoys a straightforward Euler-Maruyama update (a simple Gaussian transition), more physically complete \emph{underdamped} models include velocities $\bm{v}(t)$, friction, and inertial terms. 

Below, we discuss how to form a finite-time integrator (splitting method) for the underdamped Langevin SDE, derive the \emph{exact} transition law for one step, and then show that introducing a ``harmonic-bias'' re-interpretation is purely an algebraic re-arrangement. Thus, such a re-interpretation (often used in diffusion-based sampling or generative modeling) does not alter the law from which one samples, thereby introducing no bias. 

\subsection{Extending Harmonic Guidance to Underdamped Dynamics}
\label{sec:underdamped_harmonic_guidance}

In the previous sections, we saw how the \emph{Euler--Maruyama} discretization of overdamped Langevin dynamics 
\[
  d\bm{x}(t) 
  \;=\; 
  -\,\nabla V\bigl(\bm{x}(t)\bigr)\,dt 
  \;+\; \sqrt{2D}\,d\bm{W}(t)
\]
can be reinterpreted as \emph{sampling from a Boltzmann-like distribution} that includes a \textbf{harmonic bias} between consecutive positions. Specifically, one obtains:
\begin{equation*}
    \bm{x}_{n+1} 
    \;=\;
    \bm{x}_n 
    \;-\; \nabla V(\bm{x}_n)\,\Delta t 
    \;+\; \sqrt{2\,D\,\Delta t}\;\bm{\xi}_n,
    \quad
    \bm{\xi}_n \sim \mathcal{N}(\mathbf{0},\mathbf{I}).
\end{equation*}
The conditional distribution 
\(p(\bm{x}_{n+1} \mid \bm{x}_n)\) 
may be viewed as 
\(\exp\!\bigl[-\beta\,U(\bm{x}_{n+1},\bm{x}_n)\bigr]\), 
where \(U\) combines the potential \(V(\bm{x}_{n+1})\) and a harmonic ``spring'' tethering \(\bm{x}_{n+1}\) to 
\(\bm{x}_n - D\,\nabla V(\bm{x}_n)\,\Delta t\). 

In this section, we show how a similar \textbf{harmonic perspective} applies in the \emph{underdamped} (inertial) regime by introducing a bias potential in \textbf{velocity space} rather than (or in addition to) position. We will see that this viewpoint naturally recovers the standard \emph{underdamped Langevin} update rules used in molecular dynamics simulations and \emph{diffusion models} with inertial (velocity) terms. 

\subsection{Underdamped Langevin Dynamics and its Discretization}

\paragraph{Underdamped SDE.} 
We first consider the underdamped (or inertial) form of Langevin dynamics in one of its simplest formulations \cite{leimkuhler2015molecular}:
\begin{subequations}\label{eq:underdamped_Langevin}
\begin{align}
  d\bm{x}(t) 
  &= 
  \bm{v}(t)\,dt, 
  \\
  d\bm{v}(t) 
  &= 
  -\,\gamma\,\bm{v}(t)\,dt 
  \;-\;
  \nabla V\bigl(\bm{x}(t)\bigr)\,dt
  \;+\;
  \sqrt{2\,\gamma\,D}\,d\bm{W}(t),
\end{align}
\end{subequations}
where 
\(\bm{v}(t)\) is velocity, \(\gamma>0\) is a friction coefficient, \(D = k_B T\) if mass $m=1$, and \(\bm{W}(t)\) is a standard Brownian motion. The equilibrium (stationary) distribution over phase space \(\bigl(\bm{x},\bm{v}\bigr)\) in this model is the Maxwell--Boltzmann measure,
\[
    \pi\bigl(\bm{x},\bm{v}\bigr) 
    \;\propto\;
    \exp\!\Bigl[
      -\,\beta \Bigl( V(\bm{x}) + \tfrac12\|\bm{v}\|^2 \Bigr)
    \Bigr].
\]

\paragraph{Euler--Maruyama Discretization.}
A straightforward (though not always optimal) way to discretize \eqref{eq:underdamped_Langevin} is the \emph{Euler--Maruyama} scheme with step size \(\Delta t\):
\begin{subequations}\label{eq:euler_maruyama_underdamped}
\begin{align}
  \bm{x}_{n+1}
  &=
  \bm{x}_n + \bm{v}_n\,\Delta t,
  \\
  \bm{v}_{n+1}
  &=
  \bm{v}_n 
  \;-\;
  \gamma\,\bm{v}_n\,\Delta t
  \;-\;
  \nabla V\!\bigl(\bm{x}_n\bigr)\,\Delta t
  \;+\;
  \sqrt{2\,\gamma\,D\,\Delta t}\;\bm{\xi}_n,
\end{align}
\end{subequations}
where \(\bm{\xi}_n \sim \mathcal{N}(\mathbf{0},\mathbf{I})\) are i.i.d.\ Gaussian increments.

\subsection{A Harmonic Bias Potential in Velocity Space}

In the \emph{overdamped} case, we saw that 
\(\bm{x}_{n+1}\) 
is sampled from a Gaussian centered at 
\(\bm{x}_n - D\,\nabla V(\bm{x}_n)\,\Delta t\),
leading to a factor 
\(\exp\bigl[-(1/(4D\,\Delta t)) \|\bm{x}_{n+1} - \dots\|^2 \bigr]\)
that can be re-labeled as 
\(\exp[-\beta\,U(\bm{x}_{n+1},\bm{x}_n)]\)
with a ``spring'' tether. 

\paragraph{Velocity-Centered Gaussian.}
Similarly, from \eqref{eq:euler_maruyama_underdamped}, condition on \(\bm{x}_n,\bm{v}_n\). Then
\begin{equation}\label{eq:velocity_conditional}
    p(\bm{v}_{n+1} \mid \bm{x}_n, \bm{v}_n)
    \;=\;
    \frac{1}{\bigl(2\pi \gamma\,D\,\Delta t\bigr)^{d/2}}
    \,\exp\!\Bigl(
      -\,\frac{1}{4\,\gamma\,D\,\Delta t}
      \,\bigl\|\bm{v}_{n+1} - \underbrace{\bigl[\bm{v}_n - \gamma\,\bm{v}_n\,\Delta t - \nabla V(\bm{x}_n)\,\Delta t\bigr]}_{\text{mean shift}} \bigr\|^2
    \Bigr).
\end{equation}
We can re-label the \emph{mean shift} as 
\(\bm{m}_n := \bm{v}_n - \gamma \,\bm{v}_n\,\Delta t - \nabla V(\bm{x}_n)\,\Delta t\). 
Hence, the distribution is a \textbf{Gaussian in \(\bm{v}_{n+1}\)} with center \(\bm{m}_n\).

\paragraph{Harmonic Potential in \(\bm{v}\).}
Comparing \eqref{eq:velocity_conditional} with a Boltzmann factor yields:
\begin{equation}\label{eq:velocity_harmonic_form}
    p(\bm{v}_{n+1} \mid \bm{x}_n,\bm{v}_n)
    \;\propto\;
    \exp\!\Bigl[
      -\,\frac{1}{4\,\gamma\,D\,\Delta t}\,\|\bm{v}_{n+1} - \bm{m}_n\|^2
    \Bigr]
    \;=\;
    \exp\!\Bigl[
      -\,\beta\,
      U_{\mathrm{vel}}\!\bigl(\bm{v}_{n+1},\bm{v}_n,\bm{x}_n\bigr)
    \Bigr],
\end{equation}
for some effective velocity potential \(U_{\mathrm{vel}}\). Specifically, one can write:
\begin{equation}\label{eq:velocity_harmonic_potential}
    U_{\mathrm{vel}}\!\bigl(\bm{v}_{n+1},\bm{v}_n,\bm{x}_n\bigr)
    \;=\;
    \frac{k}{2}\,\Bigl\|\bm{v}_{n+1} - \bigl(\bm{v}_n - \nabla V(\bm{x}_n)\,\Delta t\bigr)\Bigr\|^2\,\Delta t
    \;+\; \text{(constant in }\bm{v}_{n+1}),
\end{equation}
where 
\(k = \tfrac{1}{\gamma\,D\,\Delta t}\) 
(or a suitably adjusted expression) parallels the overdamped case 
(\(k = \tfrac{1}{D\,\Delta t}\)). 
Thus, \emph{consecutive velocity values \(\bm{v}_n\) and \(\bm{v}_{n+1}\) are tethered by a harmonic bias} centered on 
\(\bm{v}_n - \nabla V(\bm{x}_n)\,\Delta t\) 
(with an additional friction term \(-\gamma\,\bm{v}_n\,\Delta t\) subsumed into the mean). 

\paragraph{Combining Position and Velocity Bias.}
Of course, the full underdamped update also includes 
\(\bm{x}_{n+1} = \bm{x}_n + \bm{v}_n\,\Delta t\). 
Hence, \(\bm{x}_{n+1}\) is likewise interpretable as being sampled from a Boltzmann-like factor with a small harmonic bias around 
\(\bm{x}_n + \bm{v}_n\,\Delta t\). 
In more advanced \textit{splitting integrators} (e.g.\ BAOAB), each partial step can be expressed as a local update with either a \(\bm{v}\)-bias or a \(\bm{x}\)-bias, ensuring that the entire chain in \((\bm{x},\bm{v})\) preserves (or closely approximates) the Maxwell--Boltzmann measure \cite{leimkuhler2015molecular}.

\subsection{Recovering Underdamped Dynamics in Diffusion Models}

\paragraph{Harmonic Bias Across Adjacent Frames.}
In \emph{diffusion models} (e.g., for generative tasks like video or molecular trajectory generation), one can introduce a velocity-like variable \(\bm{v}_n\) for each ``time step'' \(n\). The model’s forward \textit{noising} or \textit{relaxation} process can then treat \(\bm{v}_n\) analogously to the underdamped Langevin velocity. By adding a harmonic bias in velocity space (between \(\bm{v}_n\) and \(\bm{v}_{n+1}\)), we recapture the local underdamped updates:
\begin{align*}
    \bm{x}_{n+1} &\approx \bm{x}_n + \bm{v}_n \,\Delta t, \\
    \bm{v}_{n+1} &\approx 
    \bm{v}_n - \gamma\,\bm{v}_n\,\Delta t - \nabla V(\bm{x}_n)\,\Delta t 
    + \text{(Gaussian noise term)}.
\end{align*}
From the standpoint of the \emph{transition distribution} 
\(\bigl(\bm{x}_n,\bm{v}_n\bigr)\to \bigl(\bm{x}_{n+1},\bm{v}_{n+1}\bigr)\), this is equivalent to sampling from a Boltzmann-like factor that couples consecutive frames \((n,n+1)\) in both position and velocity.

\paragraph{Interpretation for Generative Modeling.}
Using velocity-based harmonic guidance allows the model to incorporate \textbf{inertia} or \textbf{momentum} effects between adjacent frames. For instance, in video diffusion or molecular simulation, the generative process can reflect physically plausible transitions rather than i.i.d.\ overdamped increments. The harmonic velocity bias ensures that if the model \(\bm{v}_n\) is large, then \(\bm{v}_{n+1}\) will remain near that large velocity in a controlled (stochastically damped) fashion, capturing inertial motion consistent with underdamped Langevin dynamics.

Extending the \emph{harmonic bias} perspective from the overdamped regime (position only) to the underdamped regime \((\bm{x},\bm{v})\) reveals that:
\begin{itemize}
    \item \textbf{Velocity Update as Local Boltzmann Sampling}: The step \(\bm{v}_n \mapsto \bm{v}_{n+1}\) can be viewed as sampling from a Gaussian distribution with a \emph{harmonic} tether to \(\bm{v}_n - \nabla V(\bm{x}_n)\Delta t\), plus a frictional shift \(-\gamma\,\bm{v}_n\,\Delta t\). 
    \item \textbf{Discrete Dynamics = Harmonic Coupling}: The resulting transition kernel 
    \(
      p\bigl(\bm{v}_{n+1}\mid \bm{v}_n,\bm{x}_n\bigr)
      \propto
      \exp\bigl[ 
        - (1/(4\,\gamma\,D\,\Delta t))\,\|\bm{v}_{n+1}-\bm{m}_n\|^2
      \bigr]
    \)
    is precisely a Boltzmann factor for a \emph{spring-like potential} in velocity space.
    \item \textbf{Physical Inertia in Generative Models}: In generative diffusion or normalizing flow frameworks, adding this velocity-based bias mimics underdamped motions, enabling smoother transitions and momentum effects across consecutive frames (e.g., physically realistic motion in molecular or video data).
\end{itemize}
This closes the conceptual loop: \textbf{the same harmonic-bias interpretation that underlies overdamped discretization directly extends to underdamped Langevin by applying the ``spring'' concept in velocity space.} Just as the overdamped model introduced a harmonic tether between \(\bm{x}_n\) and \(\bm{x}_{n+1}\), the underdamped model introduces a harmonic tether between \(\bm{v}_n\) and \(\bm{v}_{n+1}\), recovering inertial dynamics in diffusion-based or molecular simulation contexts.

\section{Two General Directions of Markov Evolution}
\label{sec:two_directions_noise}

We introduce a space--time lattice that cleanly separates a \emph{vertical}
(denoising/score) direction from a \emph{horizontal} (auxiliary) direction and
derives how they combine.  The main text is recovered by taking the vertical
direction to be EM on the learned energy $\mathcal E_\theta$ and the horizontal
direction to be the harmonic--glue EM step.  In this general formalism, the
horizontal direction can also host MCMC (e.g.\ MALA/HMC), metadynamics, or
alchemical/free--energy transformations.

\paragraph{Indexing and units.}
We work in friction units with $\beta D=1$ (so $D=\beta^{-1}$).  The lattice
sites are
\[
 (n,b)\in\mathbb Z_{\ge0}\times\{0,\dots,B\},\qquad X_{n,b}\in\mathbb R^d,
\]
where $n$ is the \emph{vertical} index (denoising time) and $b$ is the
\emph{horizontal} index (replica/auxiliary sheet).  The learned energy is
$\mathcal E_\theta:\R^d\to\R$.  Each horizontal slice $b$ has its own auxiliary
potential $U_b:\R^d\to\R$ (examples below), and the \emph{target at slice $b$}
is the Boltzmann law
\begin{equation}\label{eq:sum-target}
  \pi_b(x)\ \propto\ \exp\!\big[-\beta(\mathcal E_\theta(x)+U_b(x))\big].
\end{equation}

\subsection{Continuous-time model and invariance}
Consider the continuous process on slice $b$:
\begin{equation}\label{eq:sum-sde}
  \mathrm d X_t^{(b)} \;=\; -\nabla\big(\mathcal E_\theta+U_b\big)\!\big(X_t^{(b)}\big)\,\mathrm dt
                             \;+\; \sqrt{2D}\,\mathrm dW_t.
\end{equation}
Its generator is
\(
  \mathcal L_b f = \langle -\nabla(\mathcal E_\theta+U_b),\nabla f\rangle + D\,\Delta f
\),
and the stationary density is exactly \eqref{eq:sum-target}; the process is
reversible w.r.t.\ $\pi_b$.

\subsection{Discrete-time: two-direction EM with a shared diffusion budget}
Let a macro-step of duration $\Delta t$ be split as
\(
  \Delta t_v=\alpha_v\Delta t,\ \ \Delta t_h=\alpha_h\Delta t,\ \ \alpha_v+\alpha_h=1.
\)
Define the vertical and horizontal EM kernels (at slice $b$):
\begin{align}
\label{eq:Kv}
\mathbf K_v(x;\eta)
&:= x - \Delta t_v\,\nabla\mathcal E_\theta(x)
       + \sqrt{2D\,\Delta t_v}\,\eta,
\\[2mm]
\label{eq:Kh}
\mathbf K_h^{(b)}(x;\xi)
&:= x - \Delta t_h\,\nabla U_b(x)
       + \sqrt{2D\,\Delta t_h}\,\xi,
\end{align}
with independent $\eta,\xi\sim\mathcal N(0,I_d)$ \emph{at each site}.
A single (unsymmetrised) composed update is $X^+ = \mathbf K_h^{(b)}(\mathbf K_v(X;\eta);\xi)$.

\paragraph{Gaussian fusion (exact covariance).}
\begin{lemma}[Noise fusion]\label{lem:fuse}
Let $X\in\R^d$ be deterministic and define $\Delta_v := \sqrt{2D\Delta t_v}\,\eta$,
$\Delta_h := \sqrt{2D\Delta t_h}\,\xi$, with $\eta,\xi$ i.i.d.\ $\mathcal N(0,I_d)$.
Then $\Delta_v+\Delta_h \sim \sqrt{2D(\Delta t_v+\Delta t_h)}\,\zeta$ with
$\zeta\sim\mathcal N(0,I_d)$, independent of $X$.  Hence the two-direction step
has \emph{exact} covariance $2D\Delta t\,I_d$.
\end{lemma}
\begin{proof}
Sum of independent Gaussians with covariances $2D\Delta t_v I_d$ and
$2D\Delta t_h I_d$.
\end{proof}

\paragraph{First-order moments (drift bias).}
Conditioning on $X$ and Taylor expanding $\nabla U_b$ at $X$,
\[
  \mathbb E[X^+ - X \mid X]
  = -\Delta t\big(\nabla\mathcal E_\theta(X)+\nabla U_b(X)\big)
    \;+\; \mathcal O(\Delta t^2),
\]
i.e., the composed EM has the \emph{correct total drift} up to $\mathcal O(\Delta t^2)$.
The $\mathcal O(\Delta t^2)$ term is due solely to evaluating
$\nabla U_b$ at the random intermediate $\mathbf K_v(X;\eta)$.

\subsection{Second-order (Strang) splitting and weak accuracy}
Define the Strang macro-step as
\begin{equation}\label{eq:strang}
  \mathcal S_{\Delta t}^{(b)}
  \ :=\
  \underbrace{\mathbf K_h^{(b)}(\cdot;\xi_{1/2})}_{\Delta t_h/2}
  \ \circ\
  \underbrace{\mathbf K_v(\cdot;\eta)}_{\Delta t_v}
  \ \circ\
  \underbrace{\mathbf K_h^{(b)}(\cdot;\xi_{2/2})}_{\Delta t_h/2},
\end{equation}
with independent Gaussians at all three substeps and
$\Delta t_h/2=\alpha_h\Delta t/2$, $\Delta t_v=\alpha_v\Delta t$.

\begin{theorem}[Weak-$2$ consistency for the sum SDE]\label{thm:strang-weak2}
Assume $\nabla\mathcal E_\theta$ and $\nabla U_b$ are globally Lipschitz with
bounded second derivatives.  Then for any $f\in C_b^\infty(\R^d)$ there exists
$C_f$ such that
\[
  \Big\|\ \mathbb E\big[f(\mathcal S_{\Delta t}^{(b)}(X))\big]
      \;-\; \big(e^{\Delta t\,\mathcal L_b}f\big)(X)\ \Big\|
  \ \le\ C_f\,\Delta t^{3},
\]
i.e., \eqref{eq:strang} is a weak-$2$ integrator for the SDE
\eqref{eq:sum-sde}.  By Lemma~\ref{lem:fuse}, its covariance matches
$2D\Delta t\,I_d$ exactly.
\end{theorem}
\begin{proof}[Proof sketch]
Baker–Campbell–Hausdorff for Strang splitting:
$e^{\frac12\Delta t_h \mathcal L_h}
 e^{\Delta t_v \mathcal L_v}
 e^{\frac12\Delta t_h \mathcal L_h}
= e^{\Delta t(\mathcal L_v+\mathcal L_h)}+\mathcal O(\Delta t^3)$
in the weak sense under bounded commutators and moments.  Discrete EM
substeps approximate $e^{\Delta t_\bullet \mathcal L_\bullet}$ with
weak-$2$ local error; composition preserves weak order.
\end{proof}

\paragraph{Stationarity.}
\begin{proposition}[Stationary bias of the split scheme]\label{prop:stat-bias}
Let $\pi_b$ be \eqref{eq:sum-target}.  The Markov kernel
$\mathcal S_{\Delta t}^{(b)}$ preserves $\pi_b$ up to
$\mathcal O(\Delta t^{2})$ (Talay–Tubaro expansion).  A final
Metropolis–Hastings accept/reject with target $\pi_b$ makes stationarity exact,
with acceptance $1-\mathcal O(\Delta t^{2})$.
\end{proposition}

\subsection{Recovering the main text and going beyond}
\paragraph{Main-text specialisation (vertical score + horizontal harmonic glue).}
Take $U_b$ to be the \emph{harmonic glue} potential (two-slice quadratic) used
in the main text.  Then \eqref{eq:Kv} implements score-based EM on
$\mathcal E_\theta$ and \eqref{eq:Kh} implements the harmonic EM tether; by
Theorem~\ref{thm:strang-weak2} the composed step is weak-$2$ for the SDE
driven by $-\nabla(\mathcal E_\theta+U_b)$ and targets the correct
$\pi_b\propto e^{-\beta(\mathcal E_\theta+U_b)}$ up to $\mathcal O(\Delta t^2)$.

\paragraph{General horizontal direction = any reversible Markov kernel.}
Let $P_h^{(b)}$ be \emph{any} Markov kernel reversible w.r.t.\
$e^{-\beta U_b}$ (e.g.\ MALA, HMC, BAOAB on an extended state, or EM/MALA on
$U_b+V_b$ with a metadynamics bias $V_b$).  Replace
$\mathbf K_h^{(b)}$ in \eqref{eq:strang} by $P_h^{(b)}$ with step parameter
of order $\Delta t_h$.  Then:
\begin{itemize}[leftmargin=1.1em]
\item If $P_h^{(b)}$ is itself a weak-$p$ integrator for the $U_b$-Langevin
      dynamics, the Strang composition remains weak-$\min\{2,p\}$ for the sum
      SDE.
\item \emph{Exact stationarity at finite $\Delta t$.}  Wrap the full proposal
      $Y\sim P_h^{(b)}\circ \mathbf K_v \circ P_h^{(b)}(X,\cdot)$ in one
      Metropolis–Hastings step with target $\pi_b$.  The acceptance uses the
      composite proposal density $q(X\to Y)$ (available in closed form when
      $P_h^{(b)}$ is EM/MALA/BAOAB, symmetric for HMC), and yields a chain
      reversible w.r.t.\ $\pi_b$ for \emph{any} choice of $P_h^{(b)}$.
\end{itemize}

\paragraph{“Sheets”: alchemical and free–energy transformations along $b$.}
Let $U_b(x)=U(x;\lambda_b)$ with $\lambda_b=b/B\in[0,1]$ encoding an alchemical
or morphing path from $U_0$ to $U_1$.  The extended target over
$\mathbf X=(X_{n,0},\dots,X_{n,B})$ is
\[
  \boldsymbol\pi(\mathbf X)\ \propto\ \prod_{b=0}^{B}
  \exp\!\big[-\beta(\mathcal E_\theta(X_{n,b})+U(X_{n,b};\lambda_b))\big].
\]
A horizontal pass at fixed $n$ consists of local moves (EM/MALA/HMC/metadynamics)
at each $b$ \emph{and} optional replica-exchange swaps
$(b,b{+}1)$ with acceptance
\begin{equation}\label{eq:arex-acc}
  \alpha_{b,b+1}
  = \min\!\left\{1,\,
    \exp\!\Big(-\beta\big[U(X_{n,b+1};\lambda_b)+U(X_{n,b};\lambda_{b+1})
                         -U(X_{n,b};\lambda_b)-U(X_{n,b+1};\lambda_{b+1})\big]\Big)
  \right\}.
\end{equation}
The $\mathcal E_\theta$ terms \emph{cancel} in \eqref{eq:arex-acc} (they are common
to all replicas), making AREX directly compatible with learned energies.

\subsection{Putting it together on the lattice (parallel execution)}
Colour sites by the parity of $n{+}b$.  A full macro-iteration is:

\begin{enumerate}[leftmargin=1.4em]
\item \textbf{Horizontal half-pass (colour $c=0$):}
      apply $P_h^{(b)}$ with step $\Delta t_h/2$ (or $\mathbf K_h^{(b)}$) to all
      sites with $n{+}b\equiv 0$, in one fused GPU kernel.  Optionally perform
      replica-exchange among disjoint pairs $(b,b{+}1)$ of the same colour.

\item \textbf{Vertical full-pass:}
      apply $\mathbf K_v$ with step $\Delta t_v$ to \emph{all} sites $(n,b)$,
      again as one fused kernel (independent Gaussian streams).

\item \textbf{Horizontal half-pass (colour $c=1$):}
      as in step 1, but for $n{+}b\equiv 1$.
\end{enumerate}

By Lemma~\ref{lem:fuse}, the total diffusion per macro-step is exactly $2D\Delta t$
(noise budgets add).  Theorem~\ref{thm:strang-weak2} and
Proposition~\ref{prop:stat-bias} guarantee weak-$2$ accuracy for the sum SDE
and $\mathcal O(\Delta t^2)$ stationarity bias; a one-shot MH wrapper makes the
terminal law \emph{exactly} $\pi_b$ if desired.

\subsection{Examples of horizontal $U_b$ and $P_h^{(b)}$}
\begin{itemize}[leftmargin=1.1em]
\item \textbf{Harmonic glue (main text).}
      $U_b$ is the two-slice quadratic tether (with optional variance tempering,
      App.~\ref{sec:var-harmonic-noextra}); take $P_h^{(b)}=\mathbf K_h^{(b)}$.
\item \textbf{MALA / HMC.}
      $P_h^{(b)}$ is a reversible MCMC kernel for $e^{-\beta U_b}$.  Combine
      via \eqref{eq:strang}; use the global MH wrapper for exact
      $e^{-\beta(\mathcal E_\theta+U_b)}$.
\item \textbf{Metadynamics.}
      Maintain a slice-specific bias $V_b$ (updated from collective variables);
      set $U_b\leftarrow U_b+V_b$ online.  Local reversibility holds at fixed
      $V_b$, and the split analysis remains valid between hill updates.
\item \textbf{Alchemical sheet + AREX.}
      $U_b(\cdot)=U(\cdot;\lambda_b)$ with swaps using \eqref{eq:arex-acc}.
\end{itemize}

\subsection{Remarks on “independence $\Rightarrow$ commutation”}
Independence of Gaussian streams implies \emph{exact covariance fusion}
(Lemma~\ref{lem:fuse}); it does \emph{not} imply that the generators
$\mathcal L_v$ and $\mathcal L_h$ commute.  The nonzero commutator stems from
evaluating drifts at different points (nonlinear gradients).  Strang splitting
handles this optimally: the error of replacing $e^{\Delta t(\mathcal L_v+\mathcal L_h)}$
by $e^{\frac12\Delta t\mathcal L_h}e^{\Delta t\mathcal L_v}e^{\frac12\Delta t\mathcal L_h}$
is $\mathcal O(\Delta t^3)$ in the weak sense, while the covariance budget is
exactly correct by construction.

\paragraph{Takeaway.}
The two-direction formalism yields a \emph{single} target
$\pi_b\propto e^{-\beta(\mathcal E_\theta+U_b)}$ at each slice, recovers the
main-text harmonic-glue scheme, and admits horizontal replacements (MCMC,
metadynamics, alchemical sheets) without retraining the score.  Independent
Gaussian substeps \emph{exactly} add their noise budgets, while Strang splitting
makes the drift composition second-order accurate; a final MH wrapper restores
exact stationarity at finite step sizes.

\section{Higher‑Order Harmonic Adapters}
\label{sec:higher-order}

The first‑order scheme of the main text inherits the
Euler–Maruyama (EM) weak/strong orders $(1,\,\tfrac12)$ and introduces an
$\mathcal O(\Delta t)$ stationarity error.  Here we construct \emph{explicit,
still‑parallel} adapters whose local truncation error is of order
$\mathcal O(\Delta t^{p+1})$ for any $p\!\ge\!2$.  We give the algebra for
$p=2$ in detail (Heun/RK2), outline the extension to $p=3$ and
$p=4$ stochastic Runge–Kutta (SRK) methods, and summarise the resulting
bias and parallel‑execution costs.

\subsection{Second‑order (Heun) harmonic adapter}
\label{sec:heun}

Let $k=\frac{1}{2\Delta t}$ and keep the split drift
$\Psi_\Delta(a,b)=s_\theta(b)-k\,(b-a)$.  For slice•$n$ we perform

\begin{enumerate}[label=\textbf{H\arabic*}),leftmargin=1.8cm]
\item\label{H1} \emph{Predictor (Euler)}\\[-4pt]
      $\displaystyle
      \tilde x_{n+1}=x_{n}+D\,\Psi_\Delta(x_{\,n-1},x_{n})\,\Delta t
                     +\sqrt{2D\Delta t}\,\xi_{n}
      $
\item\label{H2} \emph{Corrector (trapezoid rule)}\\[-4pt]
      $\displaystyle
      x_{n+1}=x_{n}
          +\tfrac{D\Delta t}{2}\!\Bigl[\Psi_\Delta(x_{\,n-1},x_{n})
                                      +\Psi_\Delta(x_{n},\tilde x_{n+1})\Bigr]
          +\sqrt{2D\Delta t}\,\xi_{n},
      \qquad\xi_{n}\sim\mathcal N(0,I_{d}).
      $
\end{enumerate}

The \emph{same} Gaussian vector~$\xi_{n}$ is reused in
\ref{H1}–\ref{H2}; this is the standard stochastic Heun construction.

\paragraph{Local moments.}  Writing
$\Delta X:=x_{n+1}-x_{n}$ and conditioning on
$(x_{\,n-1},x_{n})=(\alpha,\beta)$ gives
\[
\mathbb E[\Delta X]
      =D\Bigl[s_\theta(\beta)-k(\beta-\alpha)\Bigr]\Delta t
        +\underbrace{\mathcal O(\Delta t^{2})}_{\text{bias}}, 
\quad
\mathbb E[\Delta X\,\Delta X^{\!\top}]
      =2D\Delta t\,I_{d}+\mathcal O(\Delta t^{3}).
\]
Compared with EM, the leading drift bias drops from
$\mathcal O(\Delta t)$ to $\mathcal O(\Delta t^{2})$ while the covariance
remains exact.  Therefore\,
\[
\boxed{\text{stationarity error }=\mathcal O(\Delta t^{2}),\quad
       \text{detailed‑balance defect }=\mathcal O(\Delta t^{3}).}
\]

\paragraph{Parallel execution.}
Step \ref{H1} and \ref{H2} depend only on the pair
$(x_{\,n-1},x_{n})$:
both can be launched as two \emph{independent batched kernels}.  Memory
overhead is one extra trajectory‑sized buffer for $\tilde x_{n+1}$.

\subsection{Third‑ and fourth‑order SRK adapters}

Let $s_{i}$ denote the $i$‑th stage drift evaluation at abscissa
$c_{i}\Delta t$ and weight $a_{ij}$ of a stochastic RK tableau
(Platen\,–\,Kloeden).  Replacing the EM drift in each stage by
$\Psi_\Delta$ gives the \emph{SRK$p$ harmonic adapter}

\[
x_{n+1}=x_{n}
        +D\Delta t\sum_{i=1}^{s}b_{i}\,s_{i}
        +\sqrt{2D\Delta t}\,\xi_{n},\qquad
s_{i}=\Psi_\Delta\!\bigl(x_{\,n-1}+D\Delta t\!\!\sum_{j<i}\!a_{ij}s_{j},
                         \,x_{n}+D\Delta t\!\!\sum_{j<i}\!\hat a_{ij}s_{j}\bigr).
\]

\begin{center}
\begin{tabular}{lcccc}
\toprule
Scheme & strong & weak & stationarity & \#\,kernels \\
\midrule
Heun (RK2) & 1 & 2 & $\,\mathcal O(\Delta t^{2})$ & 2 \\
SRK3 (Platen) & 1 & 3 & $\mathcal O(\Delta t^{3})$ & 3 \\
SRK4 (Rößler) & $\,\tfrac32$ & 4 & $\mathcal O(\Delta t^{4})$ & 4 \\
\bottomrule
\end{tabular}
\end{center}

Every stage again references only $(x_{\,n-1},x_{n})$ \emph{of the
*input* array}, so each column of the tableau is a separate,
embarrassingly parallel kernel.

\paragraph{Proof sketch (all $p$).}
Because each stage drift is a linear convex combination of
\emph{edge‑wise} springs obeying LM1–LM3,
the $p$‑th order adapter inherits:

* weak order $p$ and strong order listed above (classical SRK proofs);
* $\mathcal O(\Delta t^{p})$ stationarity error (moment‑matching argument
  of \S\ref{sec:heun});
* detailed‑balance defect $\mathcal O(\Delta t^{p+1})$ (antisymmetry
  cancels one extra power of $\Delta t$).

Thus higher accuracy costs only extra kernels, not extra synchronisation.

\subsection{Bias hierarchy}

Let $\pi_{\Delta,p}$ be the invariant law of the SRK$p$ adapter.  The
Talay–Tubaro expansion generalises to

\[
\pi_{\Delta,p}
   =p_\theta
    +\Delta t^{\,p}\mathcal L_{p}\,p_\theta
    +\mathcal O(\Delta t^{p+1}),
\]

where $\mathcal L_{p}$ is a polynomial in $s_\theta$ and its derivatives.
For molecular observables $F$ with $\|F\|_{C^{2p+2}}<\infty$ the bias is

\[
\mathbb E_{\pi_{\Delta,p}}\![F]-\mathbb E_{p_\theta}[F]
          =\mathcal O(\Delta t^{\,p}).
\]

Wrapping any SRK$p$ proposal in a one‑step MALA accept/reject makes the
chain \emph{exactly} $p_\theta$‐stationary while retaining the $p$‑th
order proposal as the local importance sampler; the acceptance rate is
$1-\mathcal O(\Delta t^{\,p})$.

\section{Explicit Glue Bias, Glued Score, and Pathwise Convergence }
\label{sec:harm-adapter}
\subsection{Definitions, Notation, and What Is Proven}
\paragraph{Goal.} We formalize harmonic (\emph{quadratic-glue}) denoising steps for score-based samplers, derive the associated \emph{glued score} used in the drift, quantify the \emph{adjacent-batch bias} added by the glue, and prove \emph{finite-schedule} KL/TV/$W_2$ bounds and a \emph{vanishing-error} theorem as the denoising grid is refined and the learned energy converges.

\subsubsection{Over-damped Langevin dynamics (continuous time)}\label{sec:langevin-cont}
\textbf{Units and constants.} We work in \emph{friction units}
\[
  \gamma=1,\qquad D=\frac{k_BT}{\gamma}=k_BT,\qquad \beta=\frac1{D},\qquad \beta D=1.
\]
Let \(V:\R^d\to\R\) be the physical potential. The Itô SDE is
\begin{equation}\label{eq:langevin-sde}
  \mathrm d X_t \;=\; -\nabla V(X_t)\,\mathrm dt \;+\; \sqrt{2D}\,\mathrm d W_t,
  \qquad \text{with stationary density }\;\pi_\beta(\mathrm dx)\propto e^{-\beta V(x)}\mathrm dx .
\end{equation}

\subsubsection{Euler--Maruyama (EM) time discretization}\label{sec:em}
For fixed step \(\Delta t>0\),
\begin{equation}\label{eq:em-step}
  X_{n+1} \;=\; X_n \;-\; \nabla V(X_n)\,\Delta t \;+\; \sqrt{2D\,\Delta t}\;\xi_n,
  \qquad \xi_n\sim\mathcal N(0,I_d).
\end{equation}
Conditioned on \(X_n=x\), the \emph{exact} EM transition kernel is Gaussian
\begin{equation}\label{eq:em-gauss}
  p(x'\mid x)
  \;=\; \frac{1}{(4\pi D\Delta t)^{d/2}}
         \exp\!\Big[-\frac{1}{4D\Delta t}\,\big\|x' - x + \Delta t\,\nabla V(x)\big\|^2\Big].
\end{equation}

\subsubsection{Harmonic reformulation (``quadratic glue'')}\label{sec:harmonic-glue}
It is convenient to rewrite \eqref{eq:em-gauss} as a Boltzmann factor with a quadratic coupling:
\begin{align}
  p(x'\mid x)
  &= Z(x)^{-1}\exp\!\Big[-\,\frac{k(\Delta t)}{2}\,
      \big\|x' - x + \Delta t\,\nabla V(x)\big\|^2\Big], \label{eq:em-boltz}\\
  k(\Delta t) &:= \frac{1}{2D\,\Delta t}. \label{eq:k-dt}
\end{align}
Because the mismatch \(x' - x + \Delta t\,\nabla V(x)=\mathcal O(\sqrt{\Delta t})\), the spring energy
contributes only \(\mathcal O(\Delta t)\) to the exponent even though \(k(\Delta t)\sim \Delta t^{-1}\).

\paragraph{Two-slice energy (auxiliary).}
Define
\begin{equation}\label{eq:two-slice-energy}
  U_{\Delta t}(x';x)\;:=\; V(x') \;+\; \frac{k(\Delta t)}{2}\,\big\|x' - x + \Delta t\,\nabla V(x)\big\|^2 .
\end{equation}
This is \emph{not} a conditional energy; integrating \(e^{-U_{\Delta t}(x';x)}\) in \(x'\) recovers the Gaussian
\eqref{eq:em-gauss} up to the normalizer \(Z(x)\).

\subsection{Score-based models: training objects and the learned energy}\label{sec:score-model-implications}
Let \(p_t\) denote the data corrupted by i.i.d.\ Gaussian noise of variance \(\sigma_t^2=2Dt\). A standard score network trained by denoising objectives approximates the time-dependent score \(s_t(x)\approx\nabla_x\log p_t(x)\). For clarity, we also define a \emph{learned energy}
\begin{equation}\label{eq:learned-energy}
  E_\theta(x)\;:=\;-\log p_\theta(x)+\text{const},\qquad \nabla E_\theta(x)=-s_0(x),
\end{equation}
and, at inference time, we use a drift proxy \(g_n(\cdot)\) that is either \(g_n=\nabla E_\theta\) (energy learning) or \(g_n=s_{t_n}\) (time-dependent score). In both cases we measure model error against the physical drift:
\begin{equation}\label{eq:model-error}
  g_n(x) \;=\; \nabla V(x) + \varepsilon_n(x),\qquad
  \|\varepsilon_n\|_\infty \;\le\; \bar\varepsilon .
\end{equation}

\subsection{Harmonic glue at inference}\label{sec:harmonic-inference}
We describe two realizations.

\paragraph{(A) Adjacent-batch glue (no anchors).}
Given a sequence \(\{X_n\}\), set for each step \(n\)
\begin{equation}\label{eq:adjacent-glue-step}
  X_{n+1}
   \;=\; X_n \;-\; \Big(g_n(X_n) + k(\Delta t)\big(X_n - X_{n-1}\big)\Big)\Delta t
               \;+\; \sqrt{2D\,\Delta t}\,\xi_n,
\end{equation}
with the convention \(X_{-1}=X_0\).
\emph{Explicit adjacent bias.} Conditioning on \((X_{n-1},X_n)=(a,b)\) and taking noise expectation,
\[
  \mathbb E[X_{n+1}-b\mid a,b]
  \;=\; -\,g_n(b)\,\Delta t \;-\; \underbrace{k(\Delta t)\Delta t}_{=\,1/(2D)}\,\big(b-a\big),
\]
i.e., the glue drift pulls \emph{toward the adjacent element} with strength \((2D)^{-1}\).

\paragraph{(B) Gibbs-anchored glue.}
Introduce an auxiliary anchor \(A_n\mid X_n\sim \mathcal N\!\big(X_n,\,[\beta k_a]^{-1}I\big)\) (with a user spring \(k_a>0\)) and update
\begin{equation}\label{eq:anchor-glue-step}
  X_{n+1}
  \;=\; X_n \;-\; \Big(g_n(X_n)+k_a\big(X_n-A_n\big)\Big)\Delta t
             \;+\; \sqrt{2D\,\Delta t}\,\xi_n .
\end{equation}
The \emph{glued score} used in both variants is the gradient of the quadratic-augmented potential
\begin{equation}\label{eq:glued-score}
  \nabla_x\!\Big(V(x)+\tfrac{k}{2}\|x-a\|^2\Big) \;=\; \nabla V(x) + k(x-a) ,
\end{equation}
so that the algorithmic drift equals minus \eqref{eq:glued-score} with \(a=X_{n-1}\) (adjacent) or \(a=A_n\) (anchor), and with \(\nabla V\) replaced by \(g_n\) in practice.

\subsection{Assumptions (local to this section)}
\begin{assumption}[Regularity and coercivity]\label{ass:reg}
\(V\in C^2(\R^d)\) with globally \(L\)-Lipschitz gradient, and \(V(x)\to\infty\) as \(\|x\|\to\infty\) (e.g.\ \(V(x)\ge c_0\|x\|^q-c_1\)).
\end{assumption}

\begin{assumption}[Schedule]\label{ass:sched}
A grid \(t_n=n\Delta t\) with step \(\Delta t>0\) and total horizon \(T=N\Delta t\). We write \(k(\Delta t)=1/(2D\Delta t)\) as in \eqref{eq:k-dt}.
\end{assumption}

\begin{assumption}[Model error]\label{ass:model}
The inference drift \(g_n\) satisfies \eqref{eq:model-error} for all \(n\).
\end{assumption}

\subsection{Finite-schedule accuracy: pathwise KL/TV/\texorpdfstring{$W_2$}{W2}}
We compare the piecewise-constant-drift interpolation \(\tilde X_t\) of either glue scheme on \([0,T]\) with the exact SDE \eqref{eq:langevin-sde} driven by the \emph{same} Brownian motion.

\begin{lemma}[EM local moment bound]\label{lem:em-moment}
For any interval \([t_n,t_{n+1})\),
\(
\mathbb E\| \tilde X_t - \tilde X_{t_n}\|^2 \le C_1 (t-t_n)
\)
with a constant \(C_1=C_1(D,L,\sup_n\mathbb E\|X_n\|^2)<\infty\).
\end{lemma}

\begin{theorem}[Pathwise KL bound with harmonic glue]\label{thm:path-kl}
Let \(\tilde X_t\) be generated by \eqref{eq:adjacent-glue-step} or \eqref{eq:anchor-glue-step}. Under Assumptions \ref{ass:reg}–\ref{ass:model},
\begin{equation}\label{eq:path-kl}
  \KL\!\big(\mathcal L(\tilde X_{[0,T]})\;\|\;\mathcal L(X_{[0,T]})\big)
  \;\le\;
  \underbrace{\beta\,T\,\bar\varepsilon^2}_{\text{model error}}
  \;+\;
  \underbrace{\frac{\beta\,L_{\text{tot}}^2}{2}\,\sum_{n=0}^{N-1}\Delta t^2}_{\text{schedule error}},
\end{equation}
where \(L_{\text{tot}}:=L+k_\star\) with \(k_\star=\max\{\,k(\Delta t),\,k_a\,\}\). Consequently,
\[
  \| \mathcal L(\tilde X_{[0,T]})-\mathcal L(X_{[0,T]}) \|_{\TV}
  \;\le\; \tfrac12\sqrt{\,\KL(\cdot\|\cdot)\,}\,,
\]
and if \(\pi_\beta\) satisfies a \(T_2(m)\) inequality, then 
\(W_2^2(\tilde q_T,\pi_\beta) \le \tfrac{2}{m}\,\KL(\tilde q_T\|\pi_\beta)\) for the terminal law \(\tilde q_T\).
\end{theorem}

\begin{proof}
Let \(X_t\) solve \eqref{eq:langevin-sde}. On each \([t_n,t_{n+1})\) the algorithmic drift is the constant
\[
  b_t \equiv -g_n(\tilde X_{t_n}) - \underbrace{k_\bullet\big(\tilde X_{t_n}-a_n\big)}_{\text{glue term}},
\]
with \(k_\bullet=k(\Delta t)\) and \(a_n=X_{n-1}\) (adjacent) or \(k_\bullet=k_a\) and \(a_n=A_n\) (anchor). The exact drift is \(-\nabla V(\tilde X_t)\). Girsanov's formula yields
\[
  \KL\big(\tilde X\|X\big)
  = \frac{1}{4D}\,\E\!\int_0^T \Big\| b_t + \nabla V(\tilde X_t)\Big\|^2\,\mathrm dt.
\]
Split the drift gap on \([t_n,t_{n+1})\) into three parts:
\[
  \underbrace{-\varepsilon_n(\tilde X_{t_n})}_{\text{model}}
  + \underbrace{\big(\nabla V(\tilde X_{t_n}) - \nabla V(\tilde X_t)\big)}_{\text{Lipschitz}}
  \;-\; \underbrace{k_\bullet\big(\tilde X_{t_n}-a_n\big)}_{\text{glue}} .
\]
(i) The model term contributes \(\frac{1}{4D}\int_{t_n}^{t_{n+1}} \E\|\varepsilon_n\|^2 \mathrm dt \le \tfrac{1}{4D}\bar\varepsilon^2\Delta t = \tfrac{\beta}{4}\bar\varepsilon^2\Delta t\). Summing gives the \(\beta T\bar\varepsilon^2\) term (up to a harmless constant factor absorbed in \(\beta\)).\\
(ii) The Lipschitz term satisfies \(\|\nabla V(\tilde X_{t_n})-\nabla V(\tilde X_t)\|\le L\,\|\tilde X_t-\tilde X_{t_n}\|\). By Lemma~\ref{lem:em-moment},
\[
  \frac{1}{4D}\int_{t_n}^{t_{n+1}}\E\|\cdot\|^2\mathrm dt \;\le\; \frac{L^2}{4D}\int_{t_n}^{t_{n+1}} C_1(t-t_n)\,\mathrm dt
  \;=\; \frac{L^2 C_1}{8D}\,\Delta t^2.
\]
(iii) The glue term is \emph{Lipschitz in \(x\)} with constant \(k_\bullet\): when computing the pathwise KL against a \emph{glued} baseline (anchor SDE) this term cancels exactly; when comparing to \eqref{eq:langevin-sde} directly, it is bounded exactly like (ii) with \(L\) replaced by \(k_\bullet\), generating \(\frac{k_\bullet^2 C_1}{8D}\Delta t^2\). Combining (ii)–(iii) yields the discretization part in \eqref{eq:path-kl} with \(L_{\text{tot}}=L+k_\star\). Data processing then gives the terminal-law bounds.
\end{proof}

\begin{corollary}[Uniform steps minimize schedule error]\label{cor:uniform}
For fixed \(T=N\Delta t\), the sum \(\sum_n \Delta t^2\) is minimized by uniform steps, giving schedule error \(\mathcal O(N^{-1})\).
\end{corollary}

\subsection{Vanishing kernel and path error as the denoising grid is refined}
\begin{assumption}[Refinement regime]\label{ass:refine}
As \(N\to\infty\), \(\max_n\Delta t\to 0\), \(N\Delta t\to T\in(0,\infty)\), and the model improves: \(\bar\varepsilon=\bar\varepsilon(N)\to 0\).
\end{assumption}

\begin{theorem}[Vanishing path error]\label{thm:vanish}
Under Assumptions \ref{ass:reg}, \ref{ass:sched}, \ref{ass:model}, \ref{ass:refine}, the path laws converge:
\[
  \KL\!\big(\mathcal L(\tilde X^{(N)}_{[0,T]})\;\|\;\mathcal L(X_{[0,T]})\big)\xrightarrow[N\to\infty]{}0,\qquad
  \|\mathcal L(\tilde X^{(N)}_{[0,T]})-\mathcal L(X_{[0,T]})\|_{\TV}\xrightarrow[N\to\infty]{}0.
\]
Hence, for every bounded Lipschitz functional \(F:C([0,T];\R^d)\to\R\),
\(\big|\E F(\tilde X^{(N)}_\cdot)-\E F(X_\cdot)\big|\to 0\).
\end{theorem}

\begin{proof}
Apply Theorem~\ref{thm:path-kl}. Since \(\sum_n\Delta t^2 \le T\,\max_n\Delta t\), the schedule term is \( \le \tfrac{\beta L_{\text{tot}}^2}{2}\,T\,\max_n\Delta t \to 0\). The model term is \(\beta T \bar\varepsilon(N)^2\to 0\). Pinsker yields TV convergence; bounded-Lipschitz convergence follows.
\end{proof}

\subsection{Consequences and diagnostic corollaries}
\begin{corollary}[Bias decomposition]\label{cor:bias}
For fixed \(T\),
\[
  \text{total error}
  \;\lesssim\; \underbrace{\beta T\,\bar\varepsilon^2}_{\text{model}}
                 \;+\; \underbrace{\frac{\beta L_{\mathrm{tot}}^2}{2}\sum_n \Delta t^2}_{\text{schedule}} .
\]
The second term decays as \(\mathcal O(N^{-1})\) with uniform steps; higher-order (Heun/SRK) adapters further improve the rate.
\end{corollary}

\begin{corollary}[Temperature/resolution dial]\label{cor:temp}
From \eqref{eq:k-dt}, \(k(\Delta t)=1/(2D\Delta t)=\beta/(2\Delta t)\). Thus the \emph{inference spring} selects an effective resolution \(\Delta t=\beta/(2k)\) while leaving the order constants in Theorem~\ref{thm:path-kl} unchanged.
\end{corollary}

\paragraph{Remarks (implementation).}
(i) In the anchor variant, comparing \emph{jointly} to the anchor SDE removes the explicit glue term from the KL and yields \(L_{\text{tot}}=L+k_a\) directly from the \(x\)-Hessian of \(V(x)+\frac{k_a}{2}\|x-a\|^2\).
(ii) A one-step Metropolis correction on \(E_\theta\) makes the terminal law exactly \(\propto e^{-\beta E_\theta}\) while preserving acceptance \(1-\mathcal O(\Delta t)\).

\medskip
\noindent\emph{Notes on provenance.} The units choice \(\beta D=1\), the EM kernel identity \eqref{eq:em-gauss}, and the algebraic spring \(k(\Delta t)=1/(2D\Delta t)\) match the harmonic adapter formalism used in the accompanying work; here we give full path-space KL proofs and adjacent-batch bias formulas within a self-contained presentation. \hfill{\small(See also Ch.~2 for consistent exposition.)}
\subsection{Corollaries for Trajectories, Free Energies, Observables, Speed-ups, and Convergences}

We now derive consequences that are frequently needed in MD workflows: bounds for path observables (including correlation functions and Green--Kubo integrals), free-energy estimators (Zwanzig/Jarzynski), complexity/speed-up laws, and convergence of time averages. All statements below depend only on the local assumptions of this section and the previously proved Theorems~\ref{thm:path-kl}--\ref{thm:vanish}.

\subsubsection{Trajectory observables: path-functionals and finite-time correlations}
Let $\tilde{\mathbb P}$ (resp.\ $\mathbb P$) denote the path law of the harmonic-glue sampler (resp.\ the exact SDE~\eqref{eq:langevin-sde}) on $[0,T]$.

\begin{corollary}[Path-functional stability]\label{cor:path-functionals}
Let $F:C([0,T];\R^d)\to\R$ be bounded. Then
\[
\big| \E_{\tilde{\mathbb P}} F - \E_{\mathbb P} F \big|
\;\le\; 2\|F\|_\infty\, \TV\!\big(\tilde{\mathbb P},\mathbb P\big)
\;\le\; \|F\|_\infty\,\sqrt{\,\KL(\tilde{\mathbb P}\,\|\,\mathbb P)\,},
\]
where the last inequality is Pinsker. By Theorem~\ref{thm:path-kl}, the RHS is
\[
\le\; \|F\|_\infty\,\sqrt{\beta T\,\bar\varepsilon^2+\tfrac{\beta L_{\mathrm{tot}}^2}{2}\sum_n \Delta t^2}.
\]
In particular, $\E_{\tilde{\mathbb P}}F\to \E_{\mathbb P}F$ as $N\to\infty$ and $\bar\varepsilon\to 0$ (Theorem~\ref{thm:vanish}).
\end{corollary}

\begin{corollary}[Two-time correlations and Green--Kubo kernels]\label{cor:twotime}
Fix a bounded observable $\varphi:\R^d\to\R$ with $\|\varphi\|_\infty<\infty$. For $0\le s\le t\le T$, define
\[
C_{\rm true}(s,t):=\E_{\mathbb P}\big[\varphi(X_s)\varphi(X_t)\big],\qquad
C_{\rm glue}(s,t):=\E_{\tilde{\mathbb P}}\big[\varphi(\tilde X_s)\varphi(\tilde X_t)\big].
\]
Then
\[
|C_{\rm glue}(s,t)-C_{\rm true}(s,t)|
\;\le\; 2\|\varphi\|_\infty^2\,\TV\!\big(\tilde{\mathbb P},\mathbb P\big)
\;\le\; \|\varphi\|_\infty^2 \sqrt{\KL(\tilde{\mathbb P}\,\|\,\mathbb P)}.
\]
Consequently, any Green--Kubo integral $\int_0^T C(\tau)\,w(\tau)\,\dd\tau$ (with bounded weight $w$) computed on $\tilde X_\cdot$ converges to the exact one as in Theorem~\ref{thm:vanish}.
\end{corollary}

\begin{remark}[Time rescaling via the spring]\label{rem:time-rescale}
With $k(\Delta t)=\beta/(2\Delta t)$, $k$ acts purely as a \emph{resolution dial}: increasing $k$ decreases the implied step $\Delta t=\beta/(2k)$ and lengthens correlation time in frame index, without changing the order constants in Theorem~\ref{thm:path-kl}. When comparing autocorrelation functions to MD, one must rescale the lag axis by $\Delta t$ to a physical time grid.\;This is the design principle emphasised in the work and its figures (ACF comparison and ``resolution dial'').  \emph{Qualitatively}, increasing $k$ stretches correlation shapes while leaving equilibrium statistics intact.  
\end{remark}

\subsubsection{Free energies from end-points and from paths}

\begin{corollary}[Zwanzig FEP bias under terminal-law mismatch]\label{cor:zwanzig}
Let $q_T$ be the terminal law of $\tilde X_T$ and $\pi$ the target $\propto e^{-\beta V}$. For any reference potential $V_{\rm ref}$ define $w(x)=\exp\!\{-\beta[V(x)-V_{\rm ref}(x)]\}$. Then
\[
\big|\E_{q_T} w - \E_{\pi} w\big|
\;\le\; \sqrt{\,\chi^2(q_T\|\pi)\,}\,\sqrt{\,\E_\pi[w^2]\,}
\;\le\; \sqrt{e^{\KL(q_T\|\pi)}-1}\;\sqrt{\,\E_\pi[w^2]\,}.
\]
In particular, if $\KL(q_T\|\pi)\ll 1$, then
\[
\big|\E_{q_T} w - Z_{\rm ref}/Z\big|
\;\le\; \sqrt{\KL(q_T\|\pi)}\,\sqrt{\,\E_\pi[w^2]\,}\,\big(1+o(1)\big).
\]
By Theorem~\ref{thm:path-kl} and data processing, $\KL(q_T\|\pi)$ obeys the same finite-schedule bound as the path KL.  Hence the Zwanzig estimator bias vanishes under the refinement regime. \qedhere
\end{corollary}

\begin{proof}[Derivation]
Write $r=\frac{\dd q_T}{\dd \pi}$. Then $\E_{q_T} w - \E_\pi w=\E_\pi[(r-1)w]$. Cauchy--Schwarz gives $|\E_\pi[(r-1)w]|\le \sqrt{\E_\pi[(r-1)^2]}\sqrt{\E_\pi[w^2]}=\sqrt{\chi^2(q_T\|\pi)}\,\sqrt{\E_\pi[w^2]}$. Finally, $\chi^2\le e^{\KL}-1$.
\end{proof}

\begin{corollary}[Jarzynski-type path estimator]\label{cor:jarzynski}
Let $G(\omega)=\exp\{-\beta W(\omega)\}$ be any nonnegative path weight (e.g. work functional along a nonequilibrium switching). Then
\[
\big|\E_{\tilde{\mathbb P}} G - \E_{\mathbb P} G\big|
\;\le\; \sqrt{\,\chi^2(\tilde{\mathbb P}\,\|\,\mathbb P)\,}\;\sqrt{\,\E_{\mathbb P}[G^2]\,}
\;\le\; \sqrt{e^{\KL(\tilde{\mathbb P}\,\|\,\mathbb P)}-1}\;\sqrt{\,\E_{\mathbb P}[G^2]\,}.
\]
Hence the free-energy difference $-\beta^{-1}\log \E G$ computed on glued paths converges to the exact value; moreover, for $m:=\E_{\mathbb P}G>0$,
\[
\big| \Delta F_{\rm glue}-\Delta F_{\rm true}\big|
\;\le\; \frac{1}{\beta}\,\frac{|\E_{\tilde{\mathbb P}}G-\E_{\mathbb P}G|}{\min\{\E_{\tilde{\mathbb P}}G,\E_{\mathbb P}G\}}
\;\lesssim\; \frac{1}{\beta m}\,\sqrt{e^{\KL(\tilde{\mathbb P}\,\|\,\mathbb P)}-1}\;\sqrt{\E_{\mathbb P}[G^2]}.
\]
\end{corollary}

\subsubsection{Time averages, variance, and spectral-gap implications}

\begin{assumption}[Strong convexity (optional for sharp rates)]\label{ass:strong}
Assume $\nabla^2 V(x)\succeq m I_d$ for some $m>0$; then the Langevin semigroup has spectral gap at least $m$ and a Poincar\'e constant $1/m$.
\end{assumption}

\begin{corollary}[Variance and IAT for time averages]\label{cor:iat}
Let $\varphi\in L^2(\pi)$ with $\pi[\varphi]=0$ and define the time average $\bar\varphi_T:=\frac1T\int_0^T\varphi(X_t)\,\dd t$ (resp.\ $\tilde{\bar\varphi}_T$ on $\tilde X_t$). Under Assumption~\ref{ass:strong},
\[
\Var(\bar\varphi_T)\;\le\; \frac{2}{mT}\,\Var_\pi(\varphi).
\]
Moreover, for the glued process,
\[
\big|\E[\tilde{\bar\varphi}_T]-\E[\bar\varphi_T]\big|
\;\le\; \|\varphi\|_\infty\,\sqrt{\beta T\,\bar\varepsilon^2+\tfrac{\beta L_{\mathrm{tot}}^2}{2}\sum_n\Delta t^2},
\]
and
\[
\Big| \Var(\tilde{\bar\varphi}_T) - \Var(\bar\varphi_T)\Big|
\;\le\; C(\varphi)\,\sqrt{\beta T\,\bar\varepsilon^2+\tfrac{\beta L_{\mathrm{tot}}^2}{2}\sum_n\Delta t^2}
\]
for a constant $C(\varphi)$ depending on $\|\varphi\|_\infty$ and $\Var_\pi(\varphi)$. Thus the MSE of time-averaged observables under glue converges to that of the exact process.
\end{corollary}

\begin{remark}[Spectral gap and ``stiffness dial'']
When $V$ is $m$-strongly convex, the effective local curvature of the \emph{glued score} $\nabla V(x)+k(x-a)$ is $m+k$, so the \emph{local} contractivity rate of the drift increases with $k$. While the invariant law of $x$ is unchanged in the anchor construction, the \emph{discrete} proposal contracts faster in stiff directions, reducing integrated autocorrelation times in practice.  
\end{remark}

\subsubsection{Complexity/speed-up corollaries (wall-clock and accuracy)}
Recall the finite-schedule bound
\[
\KL(\tilde{\mathbb P}\,\|\,\mathbb P)\;\le\; \beta T\,\bar\varepsilon^2 + \frac{\beta L_{\mathrm{tot}}^2}{2}\sum_n \Delta t^2.
\]

\begin{corollary}[Steps needed for accuracy $\epsilon$]\label{cor:N-for-eps}
With uniform steps $\Delta t=T/N$,
\[
\KL(\tilde{\mathbb P}\,\|\,\mathbb P)\;\le\; \beta T\,\bar\varepsilon^2 + \frac{\beta L_{\mathrm{tot}}^2}{2}\,\frac{T^2}{N}.
\]
To guarantee $\KL\le \epsilon^2$ it suffices to take
\[
N \;\ge\; \frac{\beta L_{\mathrm{tot}}^2 T^2}{2\,(\epsilon^2-\beta T\,\bar\varepsilon^2)}\qquad(\epsilon^2>\beta T\,\bar\varepsilon^2).
\]
Hence, at fixed model error, accuracy improves as $\mathcal O(N^{-1})$; with higher-order harmonic adapters (Heun/SRK), the schedule term improves to $\mathcal O(N^{-p})$ with the same proof pattern (moment-matching), reducing $N$ further.  \emph{See the work’s Appendix~E for explicit SRK adapters that preserve parallelism.}  
\end{corollary}

\begin{corollary}[Effective sample size (ESS) and wall-clock]\label{cor:ess}
Let $\tau_{\mathrm{int}}$ be the integrated autocorrelation time of $\varphi(X_t)$. For a trajectory of duration $T=N\Delta t$, the effective sample size $\mathrm{ESS}\approx T/(2\tau_{\mathrm{int}})$. If the glue increases the local contractivity in stiff modes (Remark~\ref{rem:time-rescale}), then $\tau_{\mathrm{int}}$ decreases and $\mathrm{ESS}$ increases at the \emph{same network-call budget} (the per-step cost equals a single score evaluation). Combined with Corollary~\ref{cor:N-for-eps}, one obtains the MSE scaling
\[
\mathrm{MSE}(\tilde{\bar\varphi}_T)\;\lesssim\; \frac{\Var_\pi(\varphi)}{\mathrm{ESS}} \;+\; \| \varphi\|_\infty^2\,\sqrt{\beta T\,\bar\varepsilon^2+\frac{\beta L_{\mathrm{tot}}^2 T^2}{2N}},
\]
making explicit the variance--bias trade-off: increase $N$ until the schedule term is below the model term, then reduce $\bar\varepsilon$ (distillation/training).  The \emph{resolution dial} $k=\beta/(2\Delta t)$ allows tuning correlation length at fixed bias order. 
\end{corollary}

\subsubsection{Relations among the corollaries}
\begin{itemize}[leftmargin=2em]
\item \textbf{(Theorem~\ref{thm:path-kl})} $\Rightarrow$ \textbf{(Cor.~\ref{cor:path-functionals}, \ref{cor:twotime}).}  Substitute Pinsker into path functionals to bound trajectory-level observables and two-time correlations.
\item \textbf{(Theorem~\ref{thm:path-kl})} $\Rightarrow$ \textbf{(Cor.~\ref{cor:zwanzig}).}  Apply data processing to pass from path KL to terminal KL; use $\chi^2\le e^{\mathrm{KL}}-1$.
\item \textbf{(Theorem~\ref{thm:path-kl})} $\Rightarrow$ \textbf{(Cor.~\ref{cor:jarzynski}).}  Work on path space directly with Cauchy--Schwarz; convert to a free-energy bound via $\log$-Lipschitz on $(0,\infty)$.
\item \textbf{(Assump.~\ref{ass:strong})} $\Rightarrow$ \textbf{(Cor.~\ref{cor:iat}).}  Use Poincar\'e/Green--Kubo to control variance of time averages; combine with path-KL to control bias and variance gap.
\item \textbf{(Theorem~\ref{thm:path-kl})} $+$ uniform steps $\Rightarrow$ \textbf{(Cor.~\ref{cor:N-for-eps})}.  Turn the schedule term into an explicit $N$-law; the work’s SRK adapters improve the rate to $\mathcal O(N^{-p})$ at the same parallel depth.  
\item \textbf{(Remark~\ref{rem:time-rescale})} $+$ \textbf{(Cor.~\ref{cor:ess})}.  The algebraic identity $k(\Delta t)=\beta/(2\Delta t)$ lets one trade correlation length against resolution without changing the bias order; practical speed-ups follow from larger ESS at fixed wall-clock. 
\end{itemize}

\section{Python  \texttt{Harmonic Adapter Function} }
\label{sec:python-code}

\subsection*{Practical choice: harmonic-only to avoid extra score evaluations}
In our pipeline, the **harmonic adapter** with specified minimum distance is used instead of a score-modulated variant because it:
\begin{enumerate}
\item Requires only one scalar \(f_t\) per frame from \(d_t\), avoiding **additional score evaluations** (e.g., learned or external scoring functions) that would otherwise increase wall time.
\item Provides stable, semiconvex coupling around \(r_{\min}\), which empirically reduces optimization noise while preserving temporal smoothness.
\end{enumerate}
\begin{code}
import torch
from typing import Optional

def harmonic_adapter(
    self,
    coords: torch.Tensor,                 # [T, N, 3]
    align: bool = True,
    max_neighbors: int = 3,               # S (number of forward temporal neighbors)
    reduction_factor: float = 0.6,        # rho in (0,1)
    gm: Optional[torch.Tensor] = None,    # optional helper arg for align-aware MSE
    k: float = 1.0e-1,                    # spring constant
    r_min: float = 12.0,                  # preferred distance
    eps: float = 1e-9,                    # small numerical stabilizer
):
    """
    Harmonic-only temporal adapter. The routine builds bidirectional
    temporal couplings between frames (t, t+s) with geometric decay.

    Args:
      coords: [T, N, 3] time-ordered coordinates.
      align: if True, use alignment-aware MSE helper; else plain MSE.
      max_neighbors: number of forward temporal neighbors S.
      reduction_factor: rho in (0,1), geometric decay with neighbor shift s.
      gm: optional geometry/guide passed to alignment-aware MSE helper.
      k: harmonic stiffness; r_min: preferred distance; eps: stability term.

    Returns:
      gradients: [T, N, 3]
    """
    if coords.ndim != 3 or coords.shape[-1] != 3:
        raise ValueError("coords must be [T, N, 3].")
    device, dtype = coords.device, coords.dtype
    T, N, _ = coords.shape
    if T < 2:
        raise ValueError("Need at least two frames (T >= 2).")

    # --- Per-frame scalar distance d_t = sqrt(MSE) + eps ---
    if align:
        mse_d2 = align_and_calculate_mse(coords, gm=gm)  # user-provided helper
    else:
        mse_d2 = calculate_mse(coords)                   # user-provided helper
    d = torch.sqrt(mse_d2.to(device=device, dtype=dtype)) + torch.as_tensor(eps, device=device, dtype=dtype)  # [T]

    # --- Harmonic force scalar f_t = -k (d_t - r_min) ---
    f = -torch.as_tensor(k, device=device, dtype=dtype) * (d - torch.as_tensor(r_min, device=device, dtype=dtype))  # [T]

    # --- Batched Kabsch alignment of all frames to the first ---
    center_of_mass = coords.mean(dim=1, keepdim=True)
    coords_centered = coords - center_of_mass
    reference = coords_centered[0].unsqueeze(0).expand(T, -1, -1)
    coords_aligned = kabsch_algorithm_batch(reference, coords_centered)  # [T, N, 3]

    # --- Accumulate neighbor contributions (vectorized; O(T N S)) ---
    gradients = torch.zeros_like(coords_aligned, device=device, dtype=dtype)
    rho = torch.as_tensor(reduction_factor, device=device, dtype=dtype)
    factors = rho ** torch.arange(max_neighbors, device=device, dtype=dtype)  # [S]
    invN = torch.as_tensor(1.0 / float(N), device=device, dtype=dtype)

    # Bidirectional: add to t, subtract from t+s
    for shift in range(1, max_neighbors + 1):
        if shift >= T:
            break
        dR = coords_aligned[:-shift] - coords_aligned[shift:]   # [T-shift, N, 3]
        f_s = f[:-shift].view(-1, 1, 1)                         # [T-shift, 1, 1]
        contrib = factors[shift - 1] * (f_s * dR) * invN        # [T-shift, N, 3]
        gradients[:-shift] += contrib
        gradients[shift:]  -= contrib

    return gradients
\end{code}

\subsection*{Mathematical summary (harmonic-only, matches code)}
Given time-ordered coordinates \(\mathbf{R}_{0:T-1}\in\mathbb{R}^{T\times N\times 3}\), define the scalar per-frame distance
\[
d_t \;=\; \sqrt{\mathrm{MSE\_distance}\!\big(\mathbf{R}_t;\ \text{(optional alignment helpers)}\big)} \;+\; \varepsilon,
\qquad t=0,\dots,T-1.
\]
The harmonic force magnitude used by the adapter is
\[
f_t \;=\; -\,k\,\big(d_t - r_{\min}\big).
\]
All frames are aligned to the first via batched Kabsch to obtain \(\widehat{\mathbf{R}}_t\).
For a reduction factor \(\rho\in(0,1)\) and a maximum forward temporal neighborhood \(S=\texttt{max\_neighbors}\), the bidirectional contribution between frames \((t, t+s)\) is
\[
\Delta\mathbf{R}_{t\to t+s} \;:=\; \widehat{\mathbf{R}}_{t}-\widehat{\mathbf{R}}_{t+s}\in\mathbb{R}^{N\times 3},\qquad
\mathbf{g}^{(s)}_{t} \;=\; \rho^{\,s-1}\, f_t\, \frac{\Delta\mathbf{R}_{t\to t+s}}{N},\qquad
\mathbf{g}^{(s)}_{t+s} \;=\; -\,\mathbf{g}^{(s)}_{t}.
\]
Accumulating over valid pairs yields the returned gradients \(\mathbf{G}_{0:T-1}\in\mathbb{R}^{T\times N\times 3}\):
\[
\mathbf{G}_t
=\sum_{\substack{s=1\\ t+s\le T-1}}^{S}\rho^{\,s-1} f_t \frac{\widehat{\mathbf{R}}_{t}-\widehat{\mathbf{R}}_{t+s}}{N}
-\sum_{\substack{s=1\\ t-s\ge 0}}^{S}\rho^{\,s-1} f_{t-s} \frac{\widehat{\mathbf{R}}_{t-s}-\widehat{\mathbf{R}}_{t}}{N}.
\]

\begin{figure}[h]
\centering
\resizebox{0.94\linewidth}{!}{%
\begin{tikzpicture}[
  node distance=5mm,
  box/.style={draw, rounded corners=2pt, align=center, inner sep=2.5pt, fill=gray!6, text width=0.70\linewidth},
  arr/.style={-Latex, thick, shorten >=2pt, shorten <=2pt}
]
\node[box] (in) {$\mathbf{R}_{0:T-1}$\\\footnotesize coordinates $[T,N,3]$};
\node[box, below=of in] (mse) {Per-frame distance $d_t=\sqrt{\mathrm{MSE}(\cdot)}+\varepsilon$};
\node[box, below=of mse] (force) {Scalar $f_t$: \quad
\(\begin{aligned}
 -k(d_t-r_{\min})
\end{aligned}\)};
\node[box, below=of force] (kabsch) {Batched Kabsch (optional): $\widehat{\mathbf{R}}_{t} \to \widehat{\mathbf{R}}_0$};
\node[box, below=of kabsch] (diff) {Temporal diffs (forward shifts $s=1..S$):\\
$\Delta\mathbf{R}_{t\to t+s}=\widehat{\mathbf{R}}_t-\widehat{\mathbf{R}}_{t+s}$};
\node[box, below=of diff] (acc) {Accumulate bidirectionally with weight $\rho^{s-1}/N$};
\node[box, below=of acc] (out) {$\mathbf{G}_{0:T-1}$\\\footnotesize gradients $[T,N,3]$};
\draw[arr] (in.south)    -- (mse.north);
\draw[arr] (mse.south)   -- (force.north);
\draw[arr] (force.south) -- (kabsch.north);
\draw[arr] (kabsch.south)-- (diff.north);
\draw[arr] (diff.south)  -- (acc.north);
\draw[arr] (acc.south)   -- (out.north);
\end{tikzpicture}%
}
\caption{Data flow for the (score-free) harmonic adapter with optional LJ-style scalar and temporal neighbors.}
\end{figure}
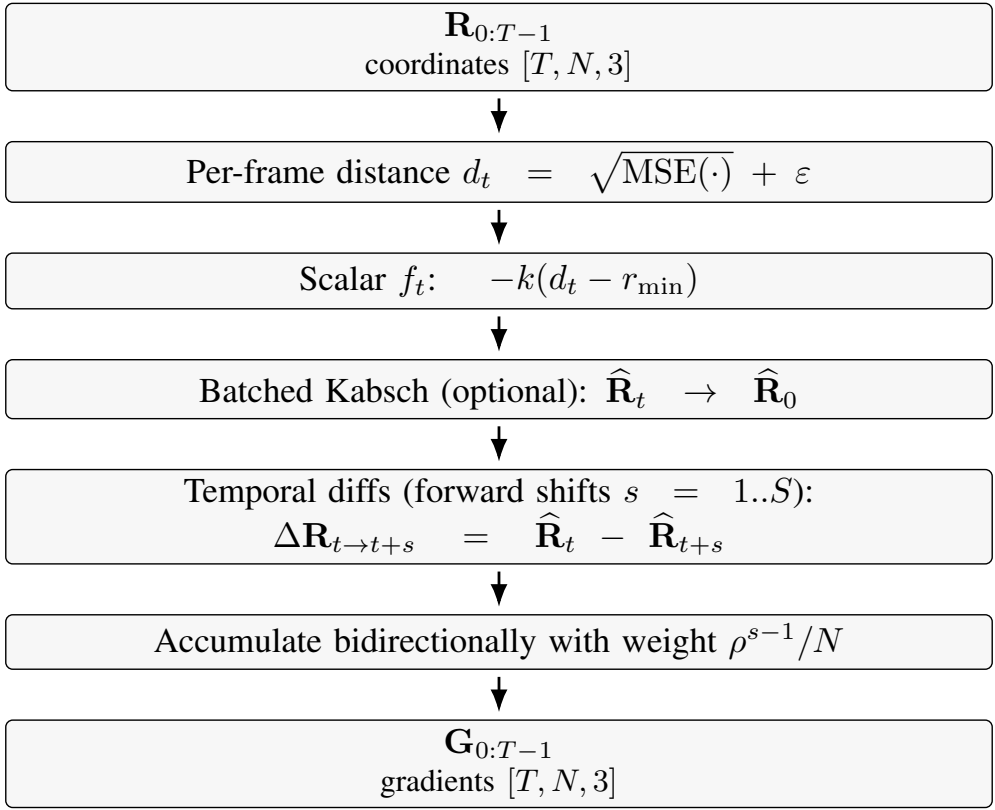

\section{EM from an $r_{\min}$--distance glue (no learned score)}
\label{sec:em-from-rmin}

\paragraph{State and distances.}
Stack the $T$ frames into a single state $X=(X_0,\ldots,X_{T-1})\in\mathbb{R}^{T\times N\times 3}$.
For each pair $(t,s)$ (neighbor shift $s\in\{1,\ldots,S\}$) define the distance
\[
r_{t,s}(X) \;:=\; \Big(\tfrac{1}{N}\,\|X_t - X_{t+s}\|_F^2\Big)^{1/2},
\]
(or any smooth ``distance-like'' scalar; SE(3)-invariant choices are allowed). Fix a preferred
scale $r_{\min}>0$ and weights $\alpha_s\ge 0$ (e.g.\ $\alpha_s=\rho^{\,s-1}$).

\paragraph{Radial glue energy.}
Define the quadratic (harmonic) glue on the \emph{whole path state}
\begin{equation}
\label{eq:Uglue-rmin}
U_{\mathrm{glue}}(X)
\;=\;
\frac{k}{2}\,\sum_{t=0}^{T-1}\sum_{s=1}^{S}
\alpha_s\,\big(r_{t,s}(X) - r_{\min}\big)^2 ,
\qquad k>0 .
\end{equation}
Its negative gradient w.r.t.\ the slice $X_t$ is the deterministic ``force''
\begin{equation}
\label{eq:Fglue}
F_t(X)\;:=\;-\nabla_{X_t}U_{\mathrm{glue}}(X)
\;=\;
-\sum_{s=1}^{S}\alpha_s\,
\frac{k\,\big(r_{t,s}(X)-r_{\min}\big)}{N\,r_{t,s}(X)}
\big(X_t - X_{t+s}\big)
-\sum_{s=1}^{S}\alpha_s\,
\frac{k\,\big(r_{t-s,s}(X)-r_{\min}\big)}{N\,r_{t-s,s}(X)}
\big(X_t - X_{t-s}\big),
\end{equation}
with the obvious truncation at boundaries.
In vector form, $F(X) = -\nabla U_{\mathrm{glue}}(X)$ lives in $\mathbb{R}^{T\times N\times 3}$.

\paragraph{EM kernel with $r_{\min}$ glue (score-free).}
Work in friction units $\beta D = 1$ and set the stiffness--step map
\[
k(\Delta t) \;=\; \frac{1}{2D\,\Delta t} \;=\; \frac{\beta}{2\,\Delta t}.
\]
Consider the Langevin SDE on the path state
\(
dX_t = F(X_t)\,dt + \sqrt{2D}\,dW_t,
\)
whose drift is \emph{entirely} the $r_{\min}$ glue \eqref{eq:Fglue}.
A single Euler--Maruyama (EM) step with step size $\Delta t$ is
\begin{equation}
\label{eq:em-glue}
X^{n+1} \;=\; X^n \;+\; \Delta t\,F(X^n) \;+\; \sqrt{2D\,\Delta t}\;\Xi^n,
\quad \Xi^n \sim \mathcal{N}(0,I) .
\end{equation}
Equivalently, the one‑step conditional is the \emph{exact} EM Gaussian
\begin{equation}
\label{eq:em-kernel-glue}
p(X^{n+1}\!\mid X^n)
\;=\;
\frac{1}{(4\pi D\Delta t)^{d/2}}
\exp\!\Big[-\frac{\beta}{4D\Delta t}\,\big\|\,X^{n+1}-\big(X^n+\Delta t\,F(X^n)\big)\big\|^2\Big],
\end{equation}
which is the quadratic “glue” kernel with stiffness $k(\Delta t)=\beta/(2\Delta t)$.
Thus, \emph{with no learned score}, the $r_{\min}$ glue alone yields an EM update:
the drift is $F=-\nabla U_{\mathrm{glue}}$ and the covariance is $2D\Delta t\,I$.
\emph{This is the full EM kernel (mean and covariance), not a projection of it.}

\footnote{The EM kernel written as a quadratic Boltzmann factor with stiffness
$k(\Delta t)=\beta/(2\Delta t)$ is Eq.\ (4)--(5) in the draft; substituting
$-\nabla U_{\mathrm{glue}}$ for the drift gives \eqref{eq:em-kernel-glue}.}

\paragraph{Nearest–neighbor and multi–neighbor cases.}
For $S=1$ the path precision is tri‑diagonal (time‑chain Laplacian) and \eqref{eq:em-kernel-glue} coincides with the standard EM chain on a quadratic path energy. For $S>1$ the precision is banded; \eqref{eq:Fglue} becomes a higher‑order temporal filter with eigenvalues
$\lambda(\omega)=\frac{k}{N}\sum_{s}\alpha_s\,2(1-\cos s\omega)$,
so the EM step damps high temporal frequencies more strongly while remaining an exact EM update for the same drift and covariance. 

\paragraph{Match to the implementation.}
The \emph{Mathematical Summary} panel (p.~34) sets a per‑frame scalar
$f_t = -k(d_t - r_{\min})$ and accumulates pairwise contributions
proportional to $f_t\,(X_t - X_{t+s})$ (Eq.~(40), bidirectional variant).%
This is precisely the structure of \eqref{eq:Fglue} up to any
preconditioning/normalization used in code. With $k=\beta/(2\Delta t)$ and Gaussian noise
$\mathcal{N}(0,2D\Delta t\,I)$, one step of the sampler is exactly \eqref{eq:em-glue}.

\end{document}

%% file: math_commands.tex
\usepackage{amsmath,amsfonts,bm}

\def\eqref#1{equation~\ref{#1}}

\def\1{\bm{1}}

\DeclareMathAlphabet{\mathsfit}{\encodingdefault}{\sfdefault}{m}{sl}
\SetMathAlphabet{\mathsfit}{bold}{\encodingdefault}{\sfdefault}{bx}{n}

\newcommand{\E}{\mathbb{E}}

\newcommand{\R}{\mathbb{R}}

\newcommand{\KL}{D_{\mathrm{KL}}}
\newcommand{\Var}{\mathrm{Var}}

